\documentclass{article}

\usepackage{arxiv}
\usepackage{natbib}

\usepackage{mathtools,amsfonts,amssymb,amsmath,amsthm}
\usepackage{hyperref}
\usepackage{url}
\usepackage{graphicx}
\usepackage{wrapfig}
\usepackage[capitalize,noabbrev]{cleveref}
\usepackage{xspace}
\usepackage{graphicx}
\usepackage{subcaption}
\usepackage{booktabs}
\usepackage{array}
\usepackage[most]{tcolorbox}

\theoremstyle{plain}

\newtheorem{proposition}{Proposition}[section]
\newtheorem{lemma}{Lemma}[section]
\usepackage{thm-restate}

\newcommand{\x}{\mathbf{x}}
\newcommand{\z}{\mathbf{z}}
\newcommand{\0}{\mathbf{0}}
\newcommand{\I}{\mathbf{I}}

\newcommand{\res}{\mathbf{r}}

\renewcommand{\v}{\mathbf{v}}

\newcommand{\pf}{\textit{PF-ODE}\xspace}
\newcommand{\pfmean}{\textit{PF-ODE $(\lambda=0)$}\xspace}
\newcommand{\vfm}{\v^\textit{FM}}
\newcommand{\vpf}{\v^{\textit{PF}}}

\newcommand{\cpf}{\pi^{\textit{PF}}}

\newcommand{\crnd}{\pi^{\textit{Random}}}

\newcommand{\cfmf}{\pi^{\textit{FM}}}

\newcommand{\crefp}[1]{(\cref{#1})}
\newcommand{\ddt}[1]{\frac{\diff #1}{\diff t}}

\newcommand{\drift}{\boldsymbol{\Delta}}

\newcommand{\Normal}{\mathcal{N}}

\newcommand{\diff}{\mathrm{d}}

\newcommand{\energy}[1]{\mathbb{E}\left\|#1\right\|^2}

\renewcommand{\headeright}{Preprint}
\renewcommand{\undertitle}{}
\date{}

\title{The Limits of Closed-Form Couplings \\ for Few-Step Generation}



\author{%
\makebox[\textwidth][c]{%
\textbf{Tobias Höppe}\textsuperscript{\mdseries 1,2,3}\footnotemark[1] \quad
\textbf{Stefan Bauer}\textsuperscript{\mdseries 1,2,3}\quad
\textbf{Qiang Liu}\textsuperscript{\mdseries 4} \quad
\textbf{Andrea Dittadi}\textsuperscript{\mdseries 1,2,3}\footnotemark[2] \quad
\textbf{Kirill Neklyudov}\textsuperscript{\mdseries 5,6,7}\footnotemark[2]}\\[0.5em]
\makebox[\textwidth][c]{%
\textsuperscript{1}\,Technical University of Munich \quad
\textsuperscript{2}\,Helmholtz AI \quad
\textsuperscript{3}\, Munich Center for Machine Learning}\\[0.1em]
\makebox[\textwidth][c]{%
\textsuperscript{4}\,University of Texas at Austin \quad
\textsuperscript{5}\,Mila -- Quebec AI Institute\quad
\textsuperscript{6}\,Universit\'e de Montr\'eal\quad
\textsuperscript{7}\,Institut Courtois}%
}

\begin{document}

\maketitle
\footnotetext[1]{Correspondence: \texttt{tobias.hoeppe@tum.de}.}
\footnotetext[2]{Equal advising.}
\vspace{-0.3in}
\enlargethispage{\baselineskip}

\begin{abstract}

    Few-step generators via flow models require complicated technical solutions such as a pretrained flow model or a consistency loss. While rectified flows have gained popularity for their success in few-step sampling, they pay a cost of needing a pre-trained model and sampling a large number of couplings per dataset. Yet for a finite dataset and a Gaussian probability path, the probability-flow vector field of the empirical distribution is available in closed form, so one could use this field as the teacher and skip pretraining altogether. We demonstrate that, surprisingly, this approach is impossible for problems of practical relevance. In particular, the analytic solution of the diffusion process in high dimensions becomes almost a linear transformation. The integration returns the noise vector drawn at initialisation, plus a drift that is empirically a faint copy of the data point, almost independent of the noise. The coupling therefore differs from an independent pairing only through this drift. As a result, training a model on these couplings is almost equivalent to training the model on the independent draws of the noise and the data, which does not provide a useful teacher. We highlight that this is the direct consequence of the high-dimensionality of the data, and that for low enough dimensions or large enough datasets, analytic couplings become more efficient teachers. We quantify this effect by the fraction of endpoint energy carried by the drift, which scales as $\frac{\log N}{d}$ for a dataset of size $N$ in a $d$-dimensional space. Finally, we provide an empirical study confirming all the theoretical results presented in the paper.
\end{abstract}

\vfill
{\centering
  \includegraphics[width=0.83\linewidth]{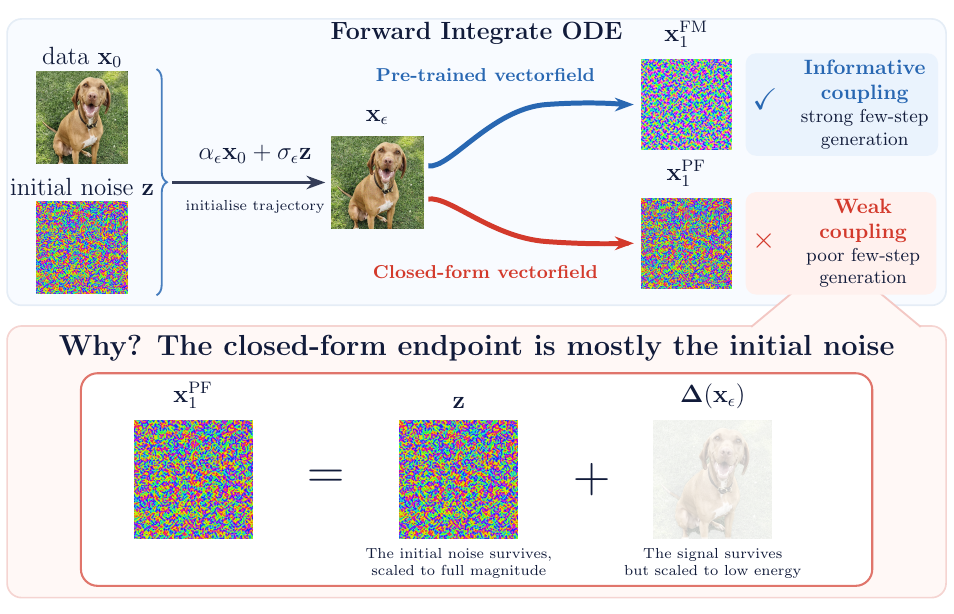}
  \label{fig:abstract}\par}
\vfill
\clearpage

\section{Introduction}

Diffusion and flow-based generative models learn a time-dependent vector field that transports samples from a simple base distribution to the data distribution by integrating an ODE \citep{sohl2015deep,ho2020denoising,songscore,lipmanflow}. Flow Matching (FM) trains a parametric field by regressing to the conditional velocity of a chosen probability path, enabling simulation-free training. Sampling, however, still requires integrating the ODE over many steps, and reducing this cost remains an active area of research.

One line of work learns a direct map between different points along the generative trajectory, enabling large jumps in time with a single model evaluation, either by distilling a pretrained model or by training from scratch \citep{salimans2022progressive,song2023consistency,albergo2025stochastic,boffi2025flow}. While distillation requires a pretrained teacher, direct training from scratch can be optimisation-sensitive and therefore requires substantial stabilisation and tuning. An alternative approach keeps the FM objective but changes the coupling between noise and data on which it trains. The regression target depends on how the endpoints are paired, and the choice of couplings can yield straighter trajectories that can be integrated in fewer steps. Several approaches construct such couplings using optimal transport \citep{pooladian2023multisample,tong2024improving,mousavi2026flow}. These couplings require no pretrained model, but the improvement they yield is limited. The most prominent approach, rectified flows \citep{liuflow}, instead obtains couplings from a pretrained teacher. In this case, each noise sample is integrated through the teacher's probability-flow ODE to generate a datapoint, and a student is trained with FM on the resulting pairs. Recent improvements have shown that a single rectification step can be enough for state-of-the-art few-step generation \citep{lee2024improving,kimsimple}. However, this procedure remains expensive, as it requires pretraining the teacher.

This raises a natural question: if rectification benefits from “good” couplings, could we avoid pretrained couplings entirely by using closed-form couplings? Concretely, given a Gaussian probability path and a dataset defining the target marginal, one can derive an “empirical” optimal regression target (vector field) for  FM/diffusion models \citep{bertrand2025closed,gao2026flow,scarvelis2025closedform}. \cite{park2026pairflow} successfully exploit this idea for discrete flow models using closed-form inversion, suggesting that a similar shortcut might be possible for rectification in the continuous case.\looseness-1

There is, however, a fundamental ambiguity in turning this vector field into couplings. The empirical probability-flow ODE maps entire regions of prior space to the same data point, so it determines which region corresponds to a data point, but not a unique endpoint within that region (\cref{fig:many_to_one}). The question is therefore whether a useful pairing can be selected directly from this closed-form flow, without relying on a pretrained teacher. We show that the natural closed-form construction produces couplings that perform close to random pairings and therefore fail to train a good student. Surprisingly, the pretrained and closed-form constructions largely associate the same data points with the same regions of prior space, but select very different endpoints within those regions. We explain this failure theoretically, identify the two quantities that control it, the dataset size $N$ and the dimension $d$, and show empirically that by moving the whole construction into lower-dimensional spaces, the closed-form couplings become better teachers.

Our contributions are as follows:
\begin{itemize}
    \item \textbf{Closed-form couplings are poor teachers} \crefp{sec:underperform}. Students trained on closed-form probability-flow couplings perform far worse than students trained on pretrained couplings, and only marginally better than students trained on random pairings. We show that this happens, even though the pretrained couplings themselves correspond to one of the pairings admitted by the closed-form field, with more than $99.9\%$ matching accuracy.
    \item \textbf{Closed-form couplings are dominated by a linear noise map} \crefp{sec:endpoint_decomposition}. We prove an endpoint decomposition for the closed-form flow (\cref{prop:endpoint_decomposition_simple}). This shows that the terminal coupling point $\x_1$ decomposes into the injected initialisation noise and a trajectory-dependent drift. We further derive an exact cutoff decomposition that bounds this drift in terms of the trajectory's tracking error (\cref{prop:drift_cutoff_simple}). Empirically, the drift is small and largely independent of the injected noise, so each datapoint is paired essentially with the fresh Gaussian sample used to initialise its integration, plus a small datapoint-dependent correction, which explains the poor performance of closed-form couplings.
    \item \textbf{Closed-form couplings require exponentially more data in higher dimensions} \crefp{sec:drift_scaling,sec:latent}. Combining our drift bound with the collapse time of empirical diffusion dynamics predicts that the informative share of the coupling scales as $\frac{\log N}{d}$. Therefore, maintaining a fixed drift share as the dimension grows requires the dataset size to scale exponentially with $d$. We test the complementary prediction by progressively reducing the image resolution. The drift share increases as $\frac{1}{d}$ and student performance improves, although the gap to pretrained couplings remains substantial even at the lowest resolutions we study.
\end{itemize}

\section{Related Work}

\paragraph{Couplings and rectified flows.} The choice of couplings used for training can significantly affect the straightness of Flow Matching trajectories. In minibatch optimal-transport methods, non-independent couplings are constructed to obtain straighter flows and therefore faster sampling \citep{pooladian2023multisample,tong2024improving}, with recent improvements by developing scalable semi-discrete variants \citep{mousavi2026flow}. Instead of relying on semi-discrete couplings, rectified flows use couplings obtained from a pre-trained teacher instead \citep{liuflow}. This idea has enabled high-quality few-step generation \citep{liu2024instaflow,lee2024improving,yan2024perflow, kimsimple}. Based on these advancements, recent theoretical studies investigate when rectification produces straight or optimal couplings \citep{bansal2026on,chambolle2026limit}. \citet{berthelot2026coupling} instead distil couplings from a pretrained normalising flow, whose endpoints are likewise obtained from noised images but, unlike in our closed-form construction, remain a learned function of the data. Since all these works require a pretrained teacher, we ask whether it can be removed altogether by obtaining endpoint pairs from the closed-form vector field. 

\paragraph{Closed-form empirical flows.} For a finite empirical distribution and Gaussian probability path, the optimal diffusion score or Flow Matching velocity can be evaluated analytically. Prior work has used these closed-form dynamics to study memorisation and generalisation \citep{scarvelis2025closedform,bertrand2025closed,gao2026flow,liu2026navigation}, including the temporal phases of the dynamics and the transition from averaging over many data points to the collapse on individual training examples \citep{biroli2024dynamical,gagneux2025generation}. Recent work further characterises the geometry produced by the exact semi-discrete Flow Matching dynamics \citep{pierret2026tessellations}. In contrast, we ask whether the endpoint coupling obtained from the empirical probability flow can itself serve as a teacher for few-step generation. We show that its datapoint-dependent information is confined to a small drift term and characterise when this signal becomes large enough to improve over random pairings.

\section{Background}
Let $\x_0 \sim p_0$ denote the data distribution on $\mathbb{R}^d$. We consider the Gaussian conditional path
\begin{equation}
\label{eq:conditional_p}
p_t(\x_t \mid \x_0) = \Normal \big(\x_t \mid \alpha_t \x_0, \sigma_t^2 \I \big)
\qquad t\in(0,1]\;,
\end{equation}
for differentiable schedules $\alpha_t, \sigma_t \in \mathbb{R}$%
\footnote{We assume $\sigma_t>0$ for $t\in(0,1]$, with $\alpha_0 = 1,\, \sigma_0 = 0$ and $\alpha_1 = 0,\, \sigma_1 = 1$.}
which induce marginals $p_t(\x)=\int p_t(\x \mid \x_0) p_0(\x_0) \diff \x_0$. A velocity field transporting these marginals can be obtained from a conditional velocity field that, together with $p_t(\x_t\mid\x_0)$, satisfies the continuity equation. For \cref{eq:conditional_p}, the conditional velocity field is 
\begin{equation}
    \v_t(\x_t \mid \x_0) = \ddt{\alpha_t}\, \x_0 + \ddt{\log \sigma_t} (\x_t - \alpha_t \x_0)\;.
\end{equation} 
The marginal probability-flow vector field is given by the posterior average $\v_t(\x) = \int \v_t(\x \mid \x_0)\, p_t(\x_0 \mid \x)\, \diff\x_0$. For an empirical distribution $\hat{p}_0(\x) = \frac{1}{N}\sum_{i=1}^{N} \delta(\x - \x_0^i)$, this integral reduces to the finite sum
\begin{equation}
\label{eq:vpf}
\vpf_t(\x) = \sum_{i=1}^N w_t^i(\x) \Big(\ddt{\alpha_t} \x_0^i + \ddt{\log \sigma_t} \big(\x - \alpha_t \x_0^i \big) \Big)\;, \qquad
w_t^i(\x)=\frac{\Normal(\x \mid \alpha_t \x_0^i, \sigma_t^2 \I)}{\sum_{j=1}^N \Normal(\x \mid \alpha_t \x_0^j, \sigma_t^2 \I)}\;.
\end{equation}

Since $\vpf_t$ is induced by the empirical data distribution, its reverse dynamics are strongly contracting as $t\rightarrow 0$. In particular, starting from a point in the Gaussian prior at $t=1$ and integrating the exact field backwards to $t=0$ sends it to one of the data points in the support of $\hat{p}_0$. Hence, whole regions of prior space can contract onto the same training example \crefp{fig:many_to_one}.  

In practice, Flow Matching approximates $\vpf_t$ with a parametric model $\vfm(\x,t)$ by least-squares regression to a Monte Carlo draw of the conditional velocity. Specifically, sample $\z \sim \Normal(\0,\I)$, set $\x_t=\alpha_t \x_0+\sigma_t \z$, and use the target $\frac{d \x_t}{dt} = \ddt{\alpha_t} \x_0 + \ddt{\sigma_t} \z$. Theoretically, a model with sufficient capacity would eventually converge to \cref{eq:vpf} \citep{yi2023generalization,kamb2024analytic}.

\begin{figure}[ht]
  \centering
  \includegraphics[width=\linewidth]{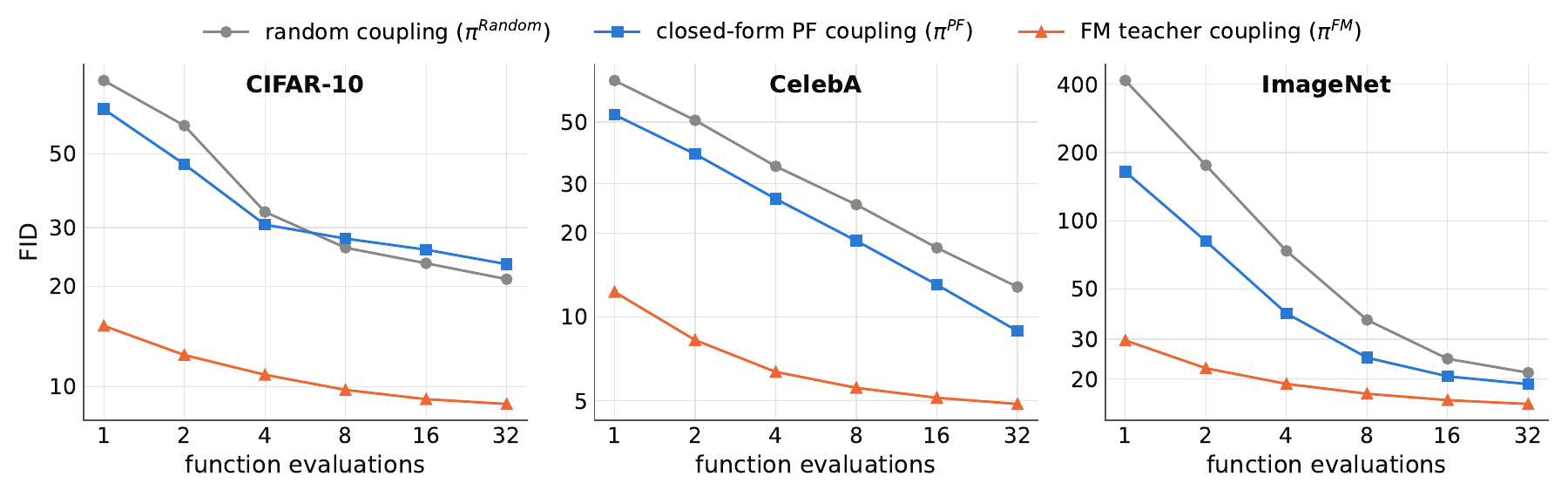}
  \caption{\textbf{FID against sampling steps for students trained on $\cpf$, $\cfmf$ and $\crnd$}. For a fair comparison, each model is trained exactly on the $N$ forward couplings permitted by the datasets. This limits the FID value, but enables us to provide a clean comparison between couplings. Although the gap narrows as the number of integration steps increases, it is substantial in the few-step regime.}
  \label{fig:cpf_vs_cfm}
\end{figure}

\section{Closed-form \pf couplings are poor teachers}
\label{sec:underperform}

A natural way to avoid pretraining a Flow Matching teacher is to construct couplings directly from the closed-form probability-flow field introduced in \cref{eq:vpf}. Surprisingly, this construction fails in practice. As shown in \cref{fig:cpf_vs_cfm}, students trained on these couplings perform much closer to students trained on random noise-data pairings than to those trained on pretrained-model couplings.  

To construct a coupling for each datapoint $\x_0^i$, we sample from the conditional path in \cref{eq:conditional_p} at a small time $\epsilon > 0$, avoiding the singularity at $t=0$:
\begin{equation}
    \label{eq:init_condition}
    \x_\epsilon = \alpha_\epsilon \x_0 + \sigma_\epsilon \z, \quad \z \sim \Normal(\0, \I)\;.
\end{equation}
Starting from $\x_\epsilon^i$, we integrate either $\vpf$ or $\vfm$ forward from $t = \epsilon$ to $t = 1$, obtaining the endpoint $\x_1^i$. We use $\epsilon=10^{-3}$ unless stated otherwise. This gives $\pi^{(\cdot)} = \{(\x_0^i,\x_1^i)\}_{i=1}^N$, with marginals $\pi^{(\cdot)}_0 = \{\x_0^i\}_{i=1}^N$ and $\pi^{(\cdot)}_1 = \{\x_1^i\}_{i=1}^N$. Here, $\cpf$ denotes couplings generated with the closed-form field and $\cfmf$ those generated with the pretrained Flow Matching model. Random couplings $\crnd$ are obtained by independently pairing datapoints with Gaussian noise samples. We use exactly $N$ forward couplings in all cases, such that the data marginal $\pi^{(\cdot)}_0$ is identical across constructions, and differences can be attributed to the induced pairing. Further details are given in \cref{sec:exp_detail}.

\cref{fig:cpf_vs_cfm} compares rectified-flow students trained with the Flow Matching objective on the three coupling sets $\cpf$, $\cfmf$, and $\crnd$. Within each dataset, all students are trained on the same number of couplings, with the same architecture and optimisation settings. The only difference is that, on CIFAR-10 and CelebA, students on $\cfmf$ are trained for fewer steps, since their FID stops improving earlier (\cref{sec:exp_detail}). The difference is most pronounced at a single sampling step, where students trained on $\cpf$ have roughly $4\text{--}5\times$ higher FID than those trained on $\cfmf$ across datasets. This gap is particularly striking because $\vpf$ is the exact probability-flow field of the empirical distribution, whereas $\vfm$ is only learned as an approximation to it. We therefore ask whether the pretrained endpoints nevertheless fall within the same regions the closed-form \pf assigns to that specific datapoint.

The large performance gap could suggest that the pretrained model learns a fundamentally different correspondence between data and prior space. To test this, we take the endpoints produced by $\vfm$ and integrate them backwards with the closed-form field $\vpf$. On all three datasets the trajectory returns to the data point from which the pretrained coupling originated in more than $99.9\%$ of the cases \crefp{app:coupling_comparison}. Thus, although $\vfm$ and $\vpf$ define different forward maps, the pretrained endpoints almost always lie within the closed-form PF-ODE region associated with the same datapoint. Combined with the observation above, this suggests that the pretrained model selects a much narrower subset within the broader regions admitted by the closed-form flow. While $\vpf$ is highly sensitive to the initial perturbation, $\vfm$ maps different perturbations of the same datapoint to much more similar endpoints \crefp{tab:coupling_narrowness}.

The results from this section isolate the source of the failure. Although the closed-form and pretrained models induce different forward maps, their endpoints are largely associated with the same datapoint regions in prior space. The difference seems to lie in how they select endpoints within these regions. The closed-form construction is highly sensitive to the initial perturbation and appears to produce a particularly uninformative pairing. In the next section, we characterise this selection explicitly and show that the endpoint is dominated by the noise used to initialise the integration.

\section{An exact endpoint decomposition: the flow returns the initialisation noise}
\label{sec:endpoint_decomposition}

Following the empirical results, we now show that the endpoint of the closed-form flow admits a simple decomposition. The integration returns the injected noise $\z$ \crefp{eq:init_condition} plus a non-linear data point dependent drift $\drift(\x_\epsilon)$, which captures how the trajectory deviates from $\x_0$ over time. This drift term is empirically small, carrying less than $0.6\%$ of the endpoint's squared norm in all datasets we investigate, and preserves the structure of the originating data point. Thus, the closed-form coupling is close to an independent noise-data pairing, with the drift $\drift(\x_\epsilon)$ providing a small datapoint-dependent correction.

We now derive this decomposition exactly. First, we split the closed-form field into a part that is linear in $\x$ and a nonlinear remainder. Using $\sum_i w_t^i(\x) = 1$, \cref{eq:vpf} can be rewritten as
\begin{equation}
\label{eq:vpf_affine}
\vpf_t(\x) = \ddt{\log\sigma_t} \x + \gamma_t m_t(\x)\;,
\quad
\gamma_t := \ddt{\alpha_t} - \alpha_t \ddt{\log\sigma_t}\;,
\quad
m_t(\x) := \sum_{i=1}^N w_t^i(\x)\,\x_0^i\;,
\end{equation}
where $m_t(\x)$ is the posterior mean of the data given $\x_t = \x$. The first term is linear in $\x$ and depends on the data only through the schedule. All nonlinearity of the field is carried by the posterior mean. The linear part can be integrated exactly, which is what produces the following identity.

\begin{restatable}[informal]{proposition}{endpointDecompositionSimple}
\label{prop:endpoint_decomposition_simple}
Let $(\x_t)_{t \in [\epsilon, 1]}$ solve $\ddt{\x_t} = \vpf_t(\x_t)$ with initial condition $\x_\epsilon = \alpha_\epsilon \x_0 + \lambda\, \sigma_\epsilon \z$, $\lambda \in [0, 1]$. Then, with $\gamma_t$ and $m_t$ as in \cref{eq:vpf_affine},

\begin{tcolorbox}[
    colback=blue!4,
    colframe=blue!35,
    boxrule=0.6pt,
    arc=2pt,
    left=2pt,
    right=2pt,
    top=2pt,
    bottom=2pt
]
\begin{equation}
\label{eq:endpoint_identity_simple}
\x_1 = \lambda \z + \drift(\x_\epsilon)\;,
\qquad
\drift(\x_\epsilon) := \int_\epsilon^1 \frac{\gamma_t}{\sigma_t}\, \big( m_t(\x_t) - \x_0 \big)\, \diff t\;.
\end{equation}
\end{tcolorbox}
\end{restatable}

\begin{figure}[ht]
  \centering
  \includegraphics[width=\linewidth]{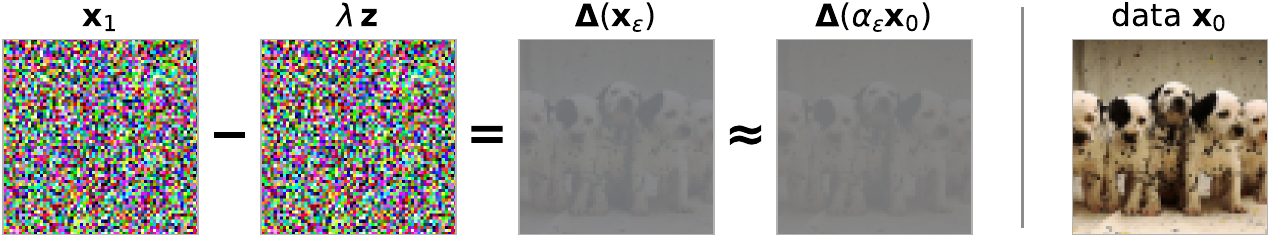}
  \caption{\textbf{Endpoint decomposition of an ImageNet-64 integration} ($\lambda = 1$, $\epsilon = 10^{-3}$,
  closed-form velocity against all $1.28$M training images). The endpoint $\x_1$ looks like pure Gaussian noise. Subtracting the injected noise $\lambda\z$ reveals the drift $\drift(\x_\epsilon)$, a faint copy of the datapoint that keeps most of its structure. The drift is well aligned with the data, with $\cos(\drift(\x_\epsilon), \x_0) = 0.88$ for this image ($0.86$ on average). The drift of the noise-free \pfmean integration, $\drift(\alpha_\epsilon\x_0)$, is almost identical to it, with $\cos(\drift(\x_\epsilon), \drift(\alpha_\epsilon\x_0)) = 0.96$ ($0.95$ on average). The injected noise therefore barely changes the information the drift carries about $\x_0$. The datapoint $\x_0$ on the right is shown for reference. All panels use the same colour scale.}
  \label{fig:drift_constant_visualisation_single}
\end{figure}

The formal proposition and its proof are given in \cref{sec:proof}. The injected noise therefore survives exactly as an additive term at the endpoint. Although it enters the initial condition only through the small perturbation $\lambda \sigma_\epsilon \z$, its endpoint contribution is $\lambda \z$, corresponding to an amplification by $\frac{1}{\sigma_\epsilon}$. For our choice $\sigma_t = t$ and $\epsilon = 10^{-3}$, this factor is $10^3$, explaining the strong sensitivity to even small amounts of initialisation noise. This exact recovery of the injected noise leaves the drift as the only component which also depends on the data point. We therefore turn to $\drift(\x_\epsilon)$ and ask what information it carries. Visually, it resembles a faint copy of the corresponding data point (\cref{fig:drift_constant_visualisation_single}, with further examples in \cref{sec:vis}), suggesting that it is largely determined by $\x_0$ rather than by the particular noise realisation. We first isolate this component by setting $\lambda=0$ and removing the noise from the integration, then test its dependence on the injected noise, and finally measure its magnitude relative to the full endpoint.

\paragraph{The case $\lambda = 0$.} Setting $\lambda = 0$  removes the injected noise and gives $\x_1 = \drift(\alpha_\epsilon \x_0)$, the drift evaluated along the mean-initialised trajectory. This provides a direct way to visualise the data point dependent part of the endpoint. As shown in \cref{fig:drift_constant_visualisation_single,fig:pf_ode_imagenet} and \cref{tab:endpoint_decomposition}(b), starting from $\x_\epsilon = \alpha_\epsilon \x_0$ produces structure-preserving endpoints rather than Gaussian noise. For datasets with reflection symmetries this is exact, as the noise-free trajectory never leaves the line through its data point, so that $\drift(\alpha_\epsilon \x_0)$ is a rescaled copy of $\x_0$ (\cref{prop:line_inv}). Therefore, while setting $\lambda = 0$ removes the injected noise that appears to make the coupling difficult to learn from, the resulting endpoints no longer have the required Gaussian marginal and cannot be used as training couplings.

\paragraph{Empirically, the drift depends only weakly on the noise.}
We test this by integrating the flow twice from every data point, with two independent noise samples $\z, \z' \sim \Normal(\0, \I)$, and subtracting the injected noise from each endpoint. This gives the two drifts $\drift(\x_\epsilon) = \x_1 - \z$ and $\drift(\x_\epsilon') = \x_1' - \z'$. A third, noise-free integration from $\x_\epsilon = \alpha_\epsilon \x_0$ gives the \pfmean endpoint $\drift(\alpha_\epsilon \x_0)$. If  the drift depended strongly on the noise, the two drifts would differ, and neither would resemble $\drift(\alpha_\epsilon \x_0)$. However, the drift is closely aligned with the \pfmean endpoint, $\cos(\drift(\x_\epsilon), \drift(\alpha_\epsilon \x_0)) = 0.94$--$0.95$, and is well described by an amplified copy of it. We found that $\drift(\x_\epsilon) \approx g\,\drift(\alpha_\epsilon \x_0)$ with $g \approx 1.2$ on all datasets, which leaves only $12$--$14\%$ of the drift energy unexplained. The drifts of the two noises agree almost as well, $\cos(\drift(\x_\epsilon), \drift(\x_\epsilon')) = 0.89$--$0.91$. They share the component $g\,\drift(\alpha_\epsilon \x_0)$ and differ only by a small noise-dependent remainder (\cref{tab:endpoint_decomposition}; see \cref{app:drift_properties} for a detailed discussion). Empirically, we therefore have the approximation
\begin{tcolorbox}[
    colback=blue!4,
    colframe=blue!35,
    boxrule=0.6pt,
    arc=2pt,
    left=2pt,
    right=2pt,
    top=2pt,
    bottom=2pt
]
\begin{equation}
\label{eq:drift_constant}
\drift(\x_\epsilon) \approx g \drift(\alpha_\epsilon \x_0)
\qquad \Longleftrightarrow \qquad
\x_1(\lambda, \epsilon) \approx g\drift(\alpha_\epsilon \x_0) + \lambda \z\;.
\end{equation}
\end{tcolorbox}
The flow therefore returns the injected noise unchanged, on top of a drift that is almost the same for every noise \crefp{fig:drift_constant_visualisation_single}.

\paragraph{Empirically, the drift is small.} Relative to the Gaussian endpoint, for which $\mathbb{E}\|\x_1\|^2=d$, it accounts for only \(0.52\%\), \(0.14\%\), and \(0.18\%\) of the squared norm on CIFAR-10, CelebA and ImageNet-64, respectively \crefp{tab:endpoint_decomposition}(a). Interestingly, the absolute drift norm remains similar across datasets despite their different dimensionalities \crefp{app:drift_properties}. 

Together with \cref{eq:drift_constant}, this gives a concrete picture of the closed-form coupling. Each data point is paired with the Gaussian sample drawn to initialise its own integration, plus a weak copy of itself, leaving only a weak datapoint-dependent signal in $\x_1$. This explains why training on $\cpf$ performs closer to training on $\crnd$ than on $\cfmf$.

\subsection{A general bound on the drift}
\label{sec:drift_cutoff}
Having established the endpoint decomposition and the empirical properties of the drift, we now ask why $\drift(\x_\epsilon)$ remains small. We derive a general bound showing that the drift is determined by how long the \pf trajectory remains close to the conditional path of its datapoint. If it remains close to this path until the signal-to-noise ratio (SNR) is already low, the drift must remain small. Empirically, this is precisely the regime we observe across all datasets, in line with the collapse behaviour identified in previous work \citep{achilli2024losing,biroli2024dynamical,gagneux2025generation,liu2026navigation}.

\begin{restatable}[informal]{proposition}{driftCutoffSimple}
\label{prop:drift_cutoff_simple}
Let $(\x_t)_{t \in [\epsilon, 1]}$ solve $\ddt{\x_t} = \vpf_t(\x_t)$ with initial condition $\x_\epsilon = \alpha_\epsilon \x_0 + \sigma_\epsilon \z$ and $\res_t := \frac{1}{\sigma_t}(\x_t - (\alpha_t \x_0 + \sigma_t \z))$ defines the normalised devitation between the conditional and unconditional trajectories. Then, for every $\tau \in (\epsilon, 1)$, the endpoint drift from \cref{eq:endpoint_identity_simple} satisfies
\begin{tcolorbox}[
    colback=blue!4,
    colframe=blue!35,
    boxrule=0.6pt,
    arc=2pt,
    left=2pt,
    right=2pt,
    top=2pt,
    bottom=2pt
]
\begin{equation}
  \|\drift(\x_\epsilon)\|_{L^2} \leq \|\res_\tau\|_{L^2} +2 \frac{\alpha_\tau}{\sigma_\tau}\sqrt{V}\;,
\end{equation}
\end{tcolorbox}
with $V$ being the total variance of the empirical data distribution and $\|\cdot\|_{L^2}$ the root-mean-square norm over $\x_0 \sim \hat p_0$ and $\z \sim \Normal(\0,\I)$.
\end{restatable}
The full proposition statement including proof can be found in \cref{sec:proof}. The bound separates the drift into a tracking-error term and an SNR-dependent term. In \cref{fig:drift_bound_evaluation} we evaluate these for all datasets on different values of $\tau$ (see \cref{app:drift_bound}). We observe that the bound is dominated by the term $2\frac{\alpha_\tau}{\sigma_\tau}\sqrt{V}$ for most of the trajectory, while the normalised tracking error $\|\res_\tau\|$ stays extremely small. This observation is consistent with previous work, which identified a regime in which the dynamics are dominated by, or even collapse onto, an individual training sample \citep{biroli2024dynamical,gagneux2025generation}. As $\tau$ increases, the trajectory starts to deviate from its conditional path while the schedule ratio $\frac{\alpha_\tau}{\sigma_\tau} \rightarrow 0$, causing the bound to become increasingly tight.

Together, the bound and our empirical results show that the drift is governed primarily by the SNR at which the trajectory stops tracking its originating data point, scaled by the overall spread of the data. The tracking error itself remains small until this transition. Understanding the drift magnitude therefore reduces to understanding when the trajectory loses track of its originating data point. In the next section, we show how this transition depends on the dataset size and dimension.

\subsection{Scaling the drift}
\label{sec:drift_scaling}

The drift is the only component of the endpoint that depends on the data point, so its significance is set by its size relative to the Gaussian noise. As the drift becomes larger relative to the Gaussian component, the datapoint-dependent signal becomes more prominent in the coupling and may therefore be easier for a student to exploit. We now ask how this ratio scales with $N$ and $d$.

\Cref{prop:drift_cutoff_simple} reduces the size of the drift to the signal-to-noise ratio $\frac{\alpha_t}{\sigma_t}$ at which the trajectory stops to track its generating datapoint. Let $\tau$ denote the timestep of this transition. As long as the trajectory remains close to its conditional path, $\x_t \approx \alpha_t \x_0^i + \sigma_t \z$, the relative posterior weight of a competing datapoint $\x_0^j$ satisfies
\begin{equation}
\log \frac{w_t^j(\x_t)}{w_t^i(\x_t)} \approx \underbrace{ \frac{\alpha_t}{\sigma_t} \z^\top(\x_0^j-\x_0^i) }_{\text{random advantage}} - \underbrace{\frac{\frac{\alpha_t}{\sigma_t}^2}{2}\|\x_0^j-\x_0^i\|^2}_{\text{distance penalty}} .
\end{equation}

In high dimensions, if pairwise distances concentrate, a typical squared distance satisfies $\|\x_0^j-\x_0^i\|^2=\Theta(V)$. Hence, conditional on the data points,
\begin{equation}
\z^\top(\x_0^j-\x_0^i) \sim \Normal\left(0, \|\x_0^j - \x_0^i\|^2 \right) \approx \Normal\left(0, V \right)\;,
\end{equation}
so a typical fluctuation has scale $\sqrt{V}$. Across $N$ competing data points, the largest such fluctuation is of order
$\sqrt{V \log N}$. The largest random advantage therefore has scale $\frac{\alpha_t}{\sigma_t} \sqrt{V \log N}$, whereas the distance penalty has scale $\frac{\alpha_t}{\sigma_t}^2V$. The posterior stops concentrating on the generating data point when these become close, i.e.
\begin{equation}
\label{eq:posterior_collapse}
\frac{\alpha_\tau}{\sigma_\tau}^2 V = \Theta(\log N)\;.
\end{equation}

\cref{eq:posterior_collapse} predicts the SNR at which the posterior loses track of the generating datapoint. We test this prediction in \cref{fig:emp_val} (left) by varying the data dimension, which in turn changes the total variance $V$, and measuring the corresponding collapse time. Across CIFAR-10, CelebA and ImageNet-64, the predicted transition closely tracks the measured one. This is the same high-dimensional collapse transition studied by \citet{biroli2024dynamical}. Their analysis identifies $\frac{\log N}{d}$ as the parameter controlling the time when the trajectory collapses to a data point and hence the start of memorisation in the reverse generative dynamics. Related analyses of the closed-form vector field observed the same temporal transition from averaging over multiple training points to dynamics dominated by a single data point \citep{bertrand2025closed,gagneux2025generation,liu2026navigation}.

Choosing $\tau$ at the beginning of this transition, the normalized tracking error $\|\res_\tau\|_{L^2}$ is empirically small \crefp{fig:drift_bound_evaluation} compared with the SNR term $\frac{\alpha_\tau}{\sigma_\tau}\sqrt V$. Hence, by \Cref{prop:drift_cutoff_simple,eq:posterior_collapse} we have
\begin{equation}
\label{eq:delta_approx}
\|\drift(\x_\epsilon)\|_{L^2} \lesssim \frac{\alpha_\tau}{\sigma_\tau} \sqrt{V} = O(\sqrt{\log N})\;.
\end{equation}
This characterisation implies that the drift grows only as $\sqrt{\log N}$. \Cref{fig:emp_val} (right) tests this prediction directly by varying the size of the empirical distribution (details in \cref{app:drift_scaling}). We found that across all three datasets, the measured drift closely follows $\mathbb{E}\|\drift(\x_\epsilon)\| \approx 1.3 \sqrt{\log N}$ up to a constant offset.

Finally, the exact probability-flow ODE preserves the prescribed marginals. Since $\alpha_1 = 0$ and $\sigma_1 = 1$, its endpoint satisfies $\x_1 \sim \Normal(\0, \I_d)$, and therefore $\mathbb E\|\x_1\|^2 = d$. The fraction of endpoint energy carried by the drift is consequently

\begin{tcolorbox}[
    colback=blue!4,
    colframe=blue!35,
    boxrule=0.6pt,
    arc=2pt,
    left=2pt,
    right=2pt,
    top=2pt,
    bottom=2pt
]
\begin{equation}
  \label{eq:drift_share}
  \frac{\mathbb E\|\drift(\x_\epsilon)\|^2}{\mathbb E\|\x_1\|^2} = O \left(\frac{\log N}{d} \right),
\end{equation}
\end{tcolorbox}

\cref{eq:drift_share} identifies the lever we could use to increase the informative fraction, as it scales linear in $\frac{1}{d}$ but only logarithmic in $N$. This motivates our next section, in which we study closed-form couplings on the same datasets for different resolutions, hence controlling $d$.

\section{Reducing the dimension improves closed-form couplings}
\label{sec:latent}

\Cref{eq:drift_share} predicts that the informative share of the closed-form coupling grows linearly in $\frac{1}{d}$ but only logarithmically in $N$. We test this by downsampling the images and repeating the coupling construction of \cref{sec:underperform}, leaving all other parts of the setup unchanged. Concretely, we vary $d = 3r^2$ through the resolution $r$ of the same images, using $r \in \{16, 32, 64\}$ for CelebA and ImageNet and $r \in \{16, 24, 32\}$ for CIFAR-10. Since the dataset size $N$ is fixed across resolutions, it does not confound the comparison. Our goal is not to reach the best possible FID, but to track how closed-form couplings perform relative to random and pretrained couplings as $d$ decreases. Full experimental details are given in \cref{sec:exp_detail} and a detailed discussion about the connection between the drift share and learnability can be found in \cref{app:learnability}.

\begin{figure}[t]
  \centering
  \includegraphics[width=\linewidth]{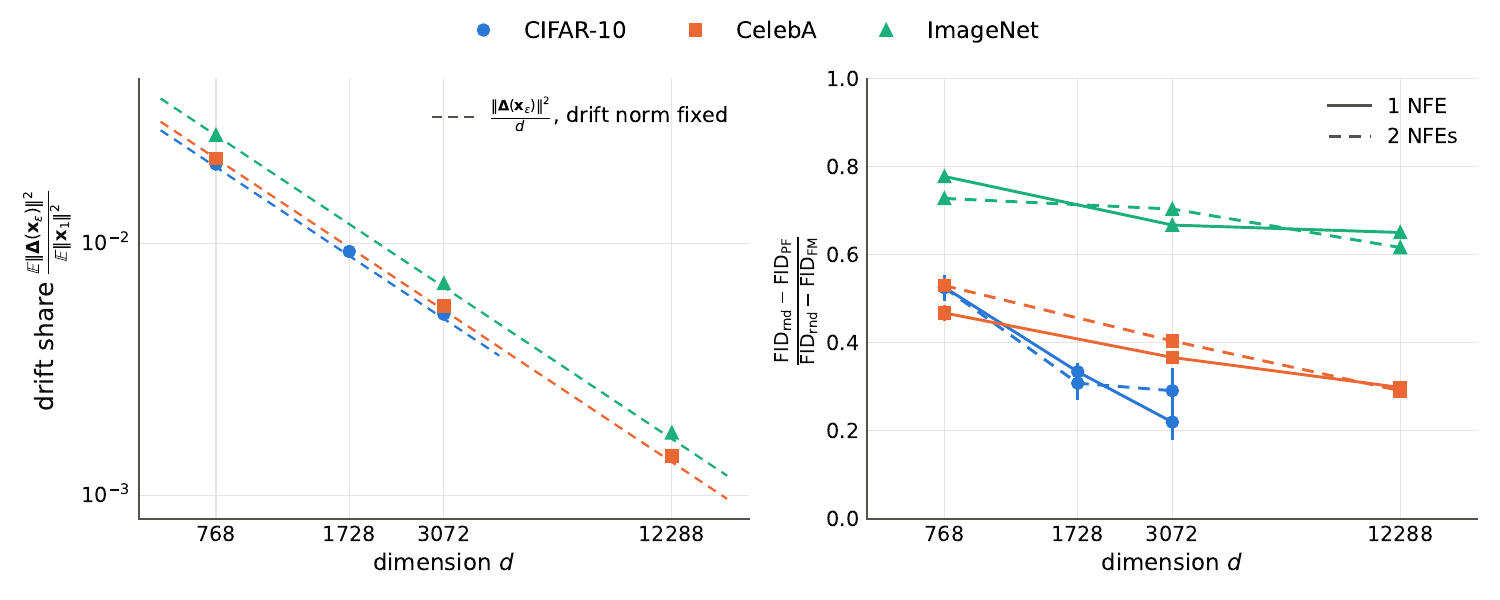}
    \caption{\textbf{Closed-form couplings against the dimension.} \textbf{Left:} Share of the endpoint's squared norm carried by the drift, $\mathbb{E}\|\drift(\x_\epsilon)\|^2 / \mathbb{E}\|\x_1\|^2$ with $\drift(\x_\epsilon) = \x_1 - \z$. The dashed lines show $\frac{\|\drift(\x_\epsilon)\|^2}{d}$ with the drift norm held fixed at its measured value. Each dataset follows this scaling closely. \textbf{Right:} Gap score $G$ \crefp{eq:relative_score} of the closed-form students at 1 and 2 Euler steps. $G = 0$ means the closed-form student is no better than the random one, and $G = 1$ means it matches the student trained on teacher couplings. $\cpf$ improves as the dimension shrinks on all datasets, but does not reach $\cfmf$. Note that $G$ is not comparable across datasets. On ImageNet, the random student cannot fit its $1.28$M random training pairs, which inflates $G$ (\cref{app:pair_fitting}).}
  \label{fig:dimension_ladder}
\end{figure}

\paragraph{Lower dimension makes the coupling more informative.} We first test the prediction of \cref{eq:drift_share} that the drift share grows linearly in $\frac{1}{d}$. As shown in \cref{fig:dimension_ladder} (left), the drift norm barely changes across resolutions, so its share of the endpoint energy grows almost exactly as $\frac{1}{d}$, from $\sim 0.15\%$ at $d = 12288$ to $\sim 2$--$2.5\%$ at $d = 768$. The same trend appears in the coupling statistics. The cosine similarity between each data point and its endpoint rises almost fourfold from $64 \times 64$ to $16 \times 16$ on CelebA and ImageNet, and doubles from $32 \times 32$ to $16 \times 16$ on CIFAR-10 \crefp{tab:cosine}. This matches the endpoint decomposition. Since $\z$ is independent of $\x_0$, we have $\cos(\x_0, \x_1) \approx \frac{\|\drift(\x_\epsilon)\|}{\sqrt{d}} \cos(\drift(\x_\epsilon), \x_0)$, which scales as $1/\sqrt{d}$ at fixed drift norm. Meanwhile, the alignment of the drift with its data point, $\cos(\drift(\x_\epsilon), \x_0)$, stays nearly constant for each dataset. The drift therefore carries similar information at every resolution, but this information becomes easier to extract as the noise shrinks relative to it. A linear readout from $\x_1$ recovers $39\%$ of the variance of held-out ImageNet images at $64 \times 64$ and $56\%$ at $16 \times 16$, with the same trend on CelebA and CIFAR-10 \crefp{tab:linear_readout}. Teacher couplings, in contrast, are strongly aligned and linearly decodable at every resolution, and even at $16 \times 16$ the closed-form couplings remain below them \crefp{tab:cosine,tab:linear_readout}.

\paragraph{More informative couplings produce better students.} We next test whether the more informative coupling translates into better students. Since FID values are not directly comparable across resolutions, we measure the gap of the closed-form student relative to the random and FM students at the same resolution,
\begin{equation}
G = \frac{\mathrm{FID}_{\mathrm{rnd}} - \mathrm{FID}_{\mathrm{PF}}}{\mathrm{FID}_{\mathrm{rnd}} - \mathrm{FID}_{\mathrm{FM}}}\ ,
\label{eq:relative_score}
\end{equation}
so that $G = 0$ corresponds to the random student and $G = 1$ to the student trained on teacher couplings. As shown in \cref{fig:dimension_ladder} (right), $G$ increases consistently as $d$ decreases (FIDs are in \cref{tab:latent_fid}). At a single step, it more than doubles on CIFAR-10 and rises by roughly $60\%$ on CelebA and $20\%$ on ImageNet. Since $G$ depends on the random baseline, only its trend within a dataset is meaningful, not its absolute level (\cref{app:pair_fitting}). Two aspects of our setup could confound the results. Downsampling changes the image content along with the dimension, and with a fixed network, the number of parameters per dimension grows as $d$ decreases. However, neither accounts for the trend. When we compress CIFAR-10 with a learned autoencoder instead of downsampling it, $G$ again grows as $d$ decreases, from $0.14$ at $d = 1024$ to $0.49$ at $d = 256$ \crefp{app:autoencoder}. Likewise, when we scale the network width to keep the number of parameters per dimension fixed, $G$ still doubles from $32$ to $16$ px \crefp{app:capacity}. A lower FID could, in principle, also come from students that merely memorise their training pairs. To test whether they actually use the drift, we construct coupling endpoints from held-out images. On these, $\cpf$ students reconstruct $33\%$–$44\%$ of the image variance on CIFAR-10, $47\%$–$61\%$ on CelebA and $40\%$–$51\%$ on ImageNet, again improving as $d$ decreases \crefp{tab:test_reconstruction}. The students therefore exploit the information in the drift rather than only their training pairs. Nor do they reproduce training images from fresh noise. Even at $16$ px, where the CIFAR-10 network has more parameters than the training set has pixel values, neither the students nor their teachers copy training images \crefp{app:memorisation}. However, the gap to the teacher couplings does not close. $\cfmf$ students reconstruct more than $92\%$ of the held-out variance at every resolution, and their FID approaches that of the teacher at low resolution \crefp{tab:teacher_fid,tab:latent_fid}. Teacher endpoints thus retain most of the information about their data point in a form that is easy to extract. In short, lowering the dimension makes closed-form couplings better teachers, but not competitive with pretrained ones.\looseness-1

\section{Conclusion}
\label{sec:conclusion}

We asked whether the couplings used for rectified flows, usually obtained from a pretrained teacher, can instead be constructed directly from the closed-form probability-flow field of the empirical data distribution. In our experiments, they cannot: students trained on these closed-form pairs perform much closer to those trained on random pairings than to those trained on teacher-generated pairs. The problem is not that the two constructions associate noise with different data points. They largely agree on which regions of noise space correspond to which data points, but choose very different endpoints within those regions. Our analytical endpoint decomposition explains why: the endpoint produced by the closed-form flow is dominated by the Gaussian noise used to initialise the trajectory, with only a small data-dependent drift. The relative contribution of this drift scales as $\frac{\log N}{d}$, so the coupling becomes less informative as the dimension grows unless the dataset size increases exponentially. We confirm empirically that closed-form couplings improve in lower-dimensional spaces. Even there, however, they remain substantially worse teachers than pretrained couplings.

This suggests that the key advantage of a learned teacher is not the coarse assignment of noise-space regions to data points, but how it selects an endpoint within each region. Understanding what determines that selection, and whether the same structure can be recovered without training a teacher, is the main open question left by our results.

\subsubsection*{Acknowledgments and Disclosure of Funding}
Part of this work was supported by the Helmholtz International Lab Causal Cell Dynamics (InterLabs-0029) through grant funding from the Initiative and Networking Fund of the Hermann von Helmholtz-Association Deutscher Forschungszentren e.V. The authors gratefully acknowledge the Gauss Centre for Supercomputing e.V. (www.gauss-centre.eu) for funding this project by providing computing time through the John von Neumann Institute for Computing (NIC) on the GCS Supercomputer JUPITER | JUWELS \citep{JUWELS} at Jülich Supercomputing Centre (JSC). Furthermore, the authors acknowledge the computational resources provided by the National High Performance Computing Centre (www.nhr.kit.edu). TH acknowledges support from G-Research. This research was enabled in part by compute resources provided by Mila (\href{https://mila.quebec/en}{mila.quebec}) and the Digital Research Alliance of Canada (\href{https://www.alliancecan.ca/en}{alliancecan.ca}). KN acknowledges support from IVADO and the Canada First Research Excellence Fund.

\bibliographystyle{unsrtnat}
\bibliography{references}

\clearpage
\appendix

\section{{Building intuition for the closed-form PF-ODE}}
\label{sec:vis}

In this section we build intuition for the closed-form field $\vpf$ and the couplings it induces. We start with a two-dimensional toy dataset, where the partition of prior space can be drawn directly and we can visualise the entire distribution. We then show that, for datasets with reflection symmetries, both this partition and the noise-free endpoints can be characterised exactly. Finally, we show that the same picture carries over to ImageNet-64.

\begin{figure}[ht]
    \centering
    \includegraphics[width=\linewidth]{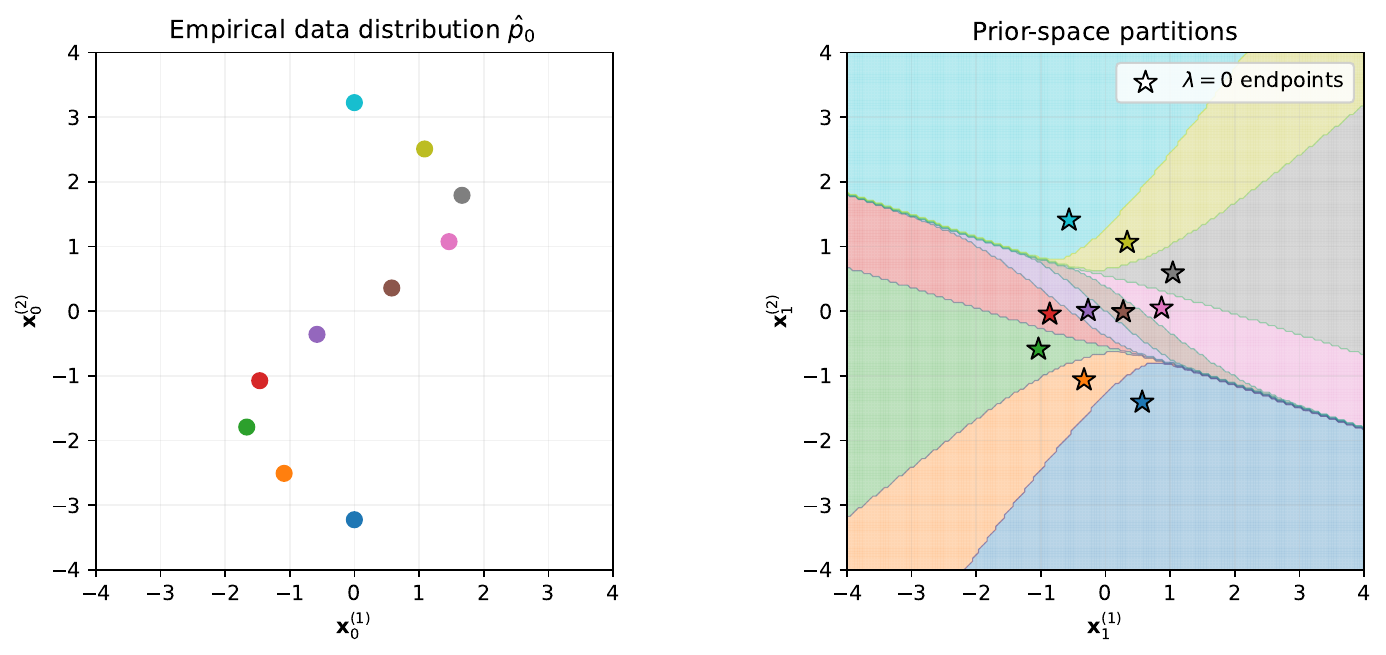}
    \caption{\textbf{Many-to-one structure of the closed-form probability-flow ODE.} \textbf{Left:} An empirical two-dimensional dataset $\hat{p}_0$, with each colour denoting one data point. \textbf{Right:} The corresponding partition of the Gaussian prior space $p_1$ induced by the closed-form \pf. Each coloured region contains all prior points that are mapped back to the same data point when the flow is integrated in reverse. The matching colours indicate the associated data point and region. Stars mark the endpoints obtained by forward integration from the conditional mean, $\x_\epsilon=\alpha_\epsilon \x_0$ ($\lambda=0$). Thus, a single data point corresponds to an entire region of prior space rather than a unique endpoint.}
    \label{fig:many_to_one}
\end{figure}

\paragraph{Many-to-one contraction.}
\Cref{fig:many_to_one} colours every point of the prior by the data point it converges to when $\vpf$ is integrated backwards from $t=1$ to $t=0$. Because the empirical dynamics contract as $t \to 0$, each data point attracts not a single point but an entire region of prior space, and together these regions tile the plane. The regions differ considerably in shape and size. Datapoints at the two ends of the arc own large regions that extend far into the tails, while those near its centre are confined to thin bands through the origin. Their probability mass under the prior, however, is identical. The exact flow transports $\Normal(\0, \I)$ onto the uniform empirical distribution, so every region carries mass $\frac{1}{N}$.

The partition also shows why the closed-form coupling is ambiguous. Every point in the region of $\x_0^i$ is a valid endpoint for $\x_0^i$, and which one we obtain is decided entirely by the initial condition. Starting from $\x_\epsilon = \alpha_\epsilon \x_0 + \sigma_\epsilon \z$, the endpoint can land anywhere in the region, depending on $\z$. Removing the noise, i.e.\ starting from $\x_\epsilon = \alpha_\epsilon \x_0$ ($\lambda = 0$), instead yields a single deterministic endpoint per data point. These endpoints trace out a contracted version of the data arc and preserve the order of the data points along it, but they cluster around the origin and no longer follow the Gaussian prior.  For symmetric datasets, both the partition and this contraction can be characterised exactly, as we show next.

\paragraph{Symmetric datasets.}
For a general dataset, the regions in \cref{fig:many_to_one} have no simple description. They depend on the global arrangement of the data. If the dataset has reflection symmetries, however, we can characterise them exactly. Both results below rely on the same observation (\cref{lem:equivariance}).  If an orthogonal map $Q$ maps the dataset onto itself, i.e. the dataset is symmetric under $Q$, then $\vpf_t$ commutes with $Q$, $\vpf_t(Q\x) = Q\,\vpf_t(\x)$. Consequently, the set of points that $Q$ leaves unchanged is invariant under the flow: a trajectory that starts there stays there. We apply this to two families of reflections. If the dataset is invariant under reflections across the bisector hyperplanes between neighbouring data points, these hyperplanes become barriers that no trajectory can cross, so each Voronoi cell of the data is mapped to its own data point (\cref{prop:voronoi_inv}). If it is invariant under reflections across the lines through individual data points, the noise-free \pfmean trajectory of each data point stays on its own line, so its endpoint is a rescaled copy of that data point (\cref{prop:line_inv}). Formal statements and proofs are given in \cref{sec:proof_symmetry}.

The first family describes the partition of prior space. If all data points have the same norm, every bisector hyperplane passes through the origin, and the Voronoi cells of the data are cones.

\begin{proposition}[Invariant Voronoi cells, informal]
\label{prop:voronoi_inv}
Let all data points have the same norm, and let $V_i$ denote the Voronoi cell of $\x_0^i$. Assume that the dataset is symmetric under the reflection across the bisector hyperplane of any two neighbouring data points, i.e. data points whose Voronoi cells share a facet. Then $\vpf_t$ is tangent to every such bisector, and a trajectory of the \pf that lies in $V_i$ at one time lies in $V_i$ at all times.
\end{proposition}

\begin{figure}[ht]
    \centering
    \includegraphics[width=\linewidth]{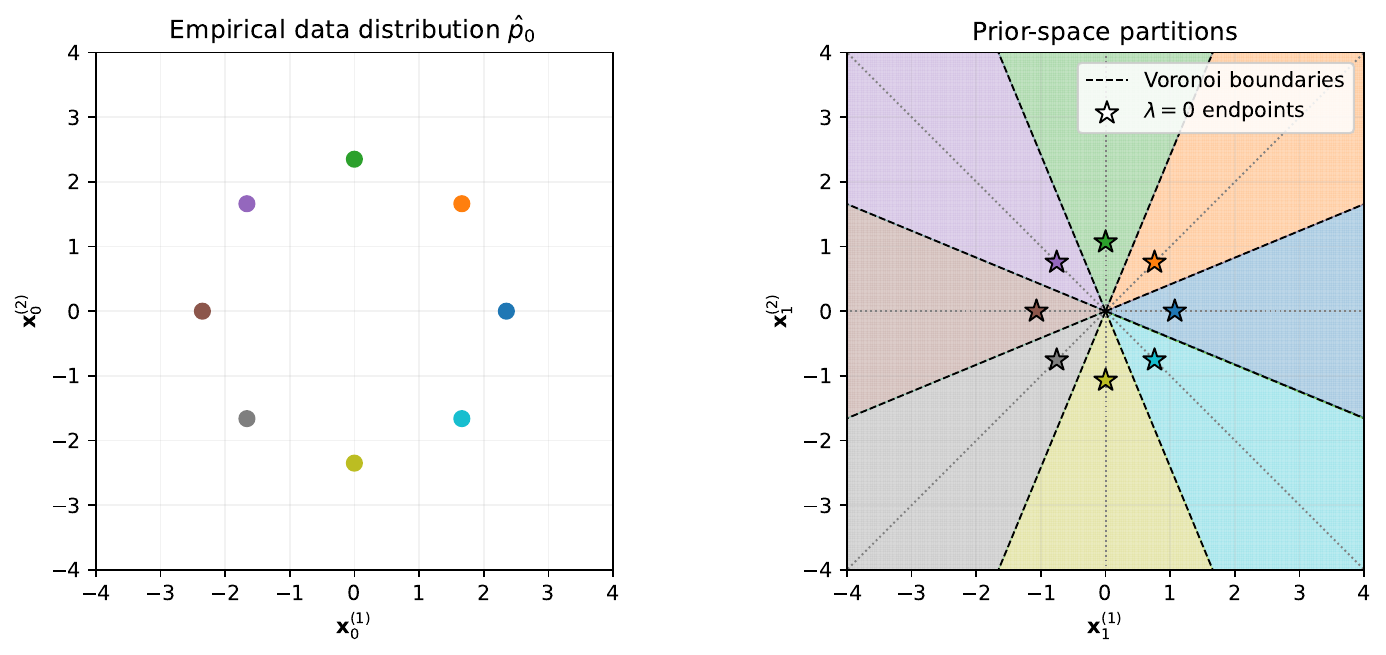}
    \caption{\textbf{Prior-space partition of a symmetric dataset.} \textbf{Left:} Eight equally spaced data points on a circle, which satisfy the assumptions of \cref{prop:voronoi_inv,prop:line_inv}. \textbf{Right:} The partition of the prior induced by the closed-form \pf, computed as in \cref{fig:many_to_one}. The regions coincide with the Voronoi cones of the data (dashed lines), as predicted by \cref{prop:voronoi_inv}. The noise-free endpoints ($\lambda = 0$, stars) lie on the rays through their data points (dotted lines), i.e.\ they are rescaled copies of the data points, as predicted by \cref{prop:line_inv}.}
    \label{fig:circle_basins}
\end{figure}

Because the cells are cones, $V_i$ is also the Voronoi cell of the conditional mean $\alpha_t \x_0^i$ at every time $t$, so the partition does not move along the path. Since the backward flow ends at a data point and $\x_0^i$ is the only data point in $V_i$, the region of prior space that the closed-form flow assigns to $\x_0^i$ is exactly its Voronoi cone. In the symmetric case, the closed-form semi-discrete coupling is therefore simply a nearest-neighbour assignment of prior samples to data points (\cref{fig:circle_basins}). In terms of \cref{prop:endpoint_decomposition_simple}, a noisy trajectory that starts in $V_i$ must also end in $V_i$, so the drift $\drift(\x_\epsilon)$ is exactly the correction that moves the injected noise $\z$ into the cell of its data point. Where in the cone the endpoint lands is still decided by $\z$, so the ambiguity of the coupling persists even in this simple geometry.

The second family describes the noise-free endpoints, marked by stars in \cref{fig:many_to_one,fig:circle_basins}.

\begin{proposition}[Invariant lines, informal]
\label{prop:line_inv}
Assume that the dataset is symmetric under the reflection across the line through a data point $\x_0^i$. Then the noise-free trajectory starting from $\x_\epsilon = \alpha_\epsilon \x_0^i$ ($\lambda = 0$) stays on this line, and its endpoint is a rescaled copy of the data point, $\drift(\alpha_\epsilon \x_0^i) = c\, \x_0^i$ for some $c \in \mathbb{R}$.
\end{proposition}

In two dimensions, this is the ordinary reflection across the line through $\x_0^i$. In higher dimensions, the required symmetry is more complex and much harder to satisfy. Both assumptions hold, for instance, for $N$ equally spaced points on a circle and for the points $\{\pm \mathbf{e}_k\}_{k=1}^d$ in $\mathbb{R}^d$. For such datasets, the closed-form flow maps the noise-free initial condition of every data point along its own ray, so the \pfmean endpoints preserve the configuration of the data up to a rescaling of each point (\cref{fig:circle_basins}). Natural images are not symmetric in this sense, and we cannot expect the statement to hold exactly. However, \cref{fig:pf_ode_imagenet,fig:drift_constant_visualisation} below show that it holds approximately. The \pfmean endpoints are faint copies of their data points, with an average cosine similarity $\cos(\drift(\alpha_\epsilon \x_0), \x_0)$ between $0.80$ and $0.93$ across datasets (\cref{tab:endpoint_decomposition}).

\begin{figure}[ht]
  \centering
  \begin{subfigure}[b]{0.45\textwidth}
    \centering
    \includegraphics[width=\linewidth]{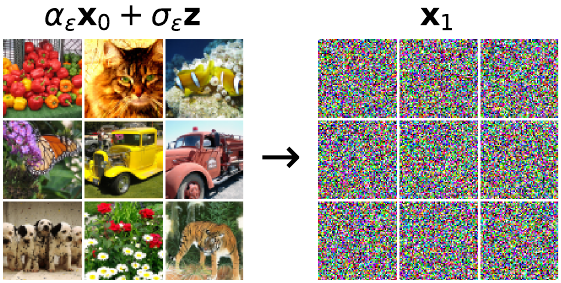}
    \caption{\textbf{Integrating $\vpf$ from $\alpha_\epsilon \x_0 + \sigma_\epsilon \z$.}}
    \label{fig:pf-ode_imagenet}
  \end{subfigure}
  \hfill
  \begin{subfigure}[b]{0.45\textwidth}
    \centering
    \includegraphics[width=\linewidth]{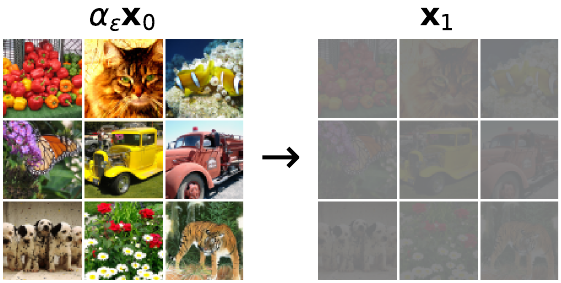}
    \caption{\textbf{Integrating $\vpf$ from $\alpha_\epsilon \x_0$.}}
    \label{fig:pf-mean_imagenet}
  \end{subfigure}
  \caption{\textbf{Noisy versus noise-free initialisation on ImageNet-64} ($\epsilon = 10^{-3}$). The two initial conditions differ only by $\sigma_\epsilon \z$ and are visually indistinguishable. Their endpoints, however, differ completely. Starting from the noisy initialisation \textbf{(a)} produces endpoints that resemble Gaussian noise, whereas starting from the conditional mean \textbf{(b)} produces faint, low-contrast copies of the data points, from which each pair can be identified by eye.}
  \label{fig:pf_ode_imagenet}
\end{figure}

\paragraph{Noisy versus noise-free initialisation.}
\Cref{fig:pf_ode_imagenet} shows that the two initialisations behave analogously on ImageNet-64. We integrate $\vpf$ forward either from the noisy initial condition $\x_\epsilon = \alpha_\epsilon \x_0 + \sigma_\epsilon \z$ used to construct $\cpf$, or from the noise-free initial condition $\x_\epsilon = \alpha_\epsilon \x_0$ used by \pfmean. Note that the latter is the mean of the conditional path in \cref{eq:conditional_p} at time $\epsilon$. Although the two initial conditions are visually indistinguishable, their endpoints differ enormously. The noisy initialisation produces endpoints that look like pure Gaussian noise, with no visible trace of the image. The noise-free initialisation produces a scaled down version of the image that is easily matched to its data point. This is the high-dimensional counterpart of \cref{fig:many_to_one}. The injected noise decides where within its region the endpoint lands and, in doing so, hides the data point, while the noise-free endpoint retains the structure of the data point but is far too small in norm to be a sample from the prior.

\begin{figure}[ht]
  \centering
  \includegraphics[width=\linewidth]{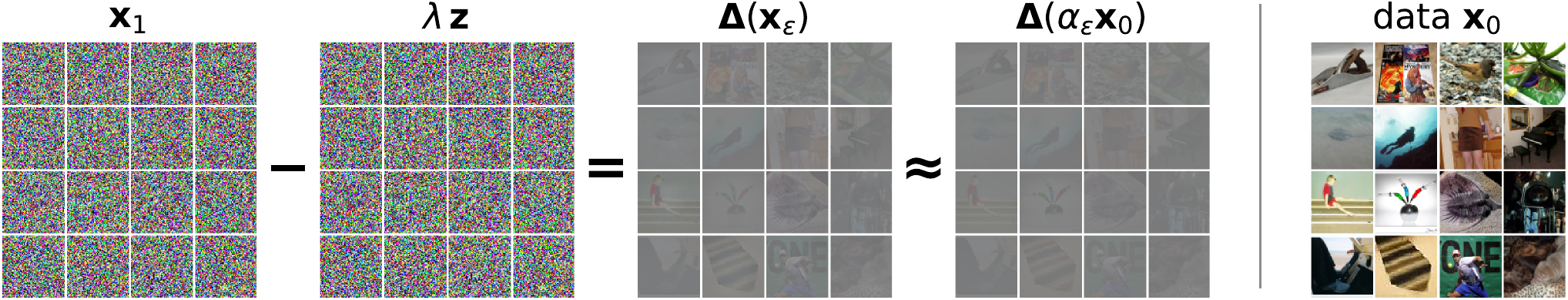}
  \caption{\textbf{Visualising the endpoint decomposition on ImageNet-64} ($\lambda = 1$, $\epsilon = 10^{-3}$). The endpoints $\x_1$ resemble pure Gaussian noise, yet subtracting the injected noise $\lambda \z$ reveals the drift $\drift(\x_\epsilon)$, a faint, structure-preserving copy of each data point that closely matches the \pfmean endpoint $\drift(\alpha_\epsilon \x_0)$ of a noise-free integration. The two drifts differ mainly in magnitude, with $\drift(\x_\epsilon) \approx 1.2\, \drift(\alpha_\epsilon \x_0)$, while their directions agree closely, with an average cosine similarity of $0.95$ (\cref{tab:endpoint_decomposition}).}
  \label{fig:drift_constant_visualisation}
\end{figure}

\paragraph{Recovering the drift.}
Finally, \cref{fig:drift_constant_visualisation} connects the two initialisations through the endpoint decomposition of \cref{prop:endpoint_decomposition_simple}. Since $\x_1 = \lambda \z + \drift(\x_\epsilon)$ holds exactly, subtracting the injected noise from the Gaussian-looking endpoints removes precisely the component that hides the image. What remains is the drift $\drift(\x_\epsilon)$, again a faint copy of the data point, and nearly indistinguishable from the \pfmean endpoint $\drift(\alpha_\epsilon \x_0)$. The information about $\x_0$ is therefore not lost during the noisy integration. It is merely buried under the injected noise, which carries more than $99\%$ of the endpoint energy. Moreover, the drift looks almost the same whether or not noise was injected, which is the approximation stated in \cref{eq:drift_constant}. This is a high-dimensional effect. In two dimensions, the drift must carry every endpoint back into the region of its data point (\cref{fig:many_to_one}, and exactly into its Voronoi cone for symmetric data, \cref{prop:voronoi_inv}) and therefore depends more strongly on the noise, and even across our image resolutions, $\cos(\drift(\x_\epsilon), \x_0)$ drops slightly as the dimension decreases \crefp{tab:cosine}. We quantify these observations in \cref{app:drift_properties}.

\section{Comparing teacher and closed-form couplings}
\label{app:coupling_comparison}

In this section, we provide the details for the two observations from \cref{sec:underperform} that compare teacher and closed-form couplings. First, the teacher couplings are consistent with the closed-form field, since teacher endpoints lie in the closed-form region of their own data point. Second, within this region, the teacher selects a much narrower set of endpoints than the closed-form flow.

We take $10000$ random teacher couplings $(\x_0^i, \x_1^i)$ per dataset and integrate each endpoint $\x_1^i$ backwards through the closed-form field of the full training set, from $t = 1$ to $t = 10^{-3}$. A coupling counts as a match if the nearest training image of the result is $\x_0^i$, which by chance happens with probability $\frac{1}{N}$. The matching accuracy is $99.95\%$ on CIFAR-10, $99.94\%$ on CelebA and $100\%$ on ImageNet. Almost every teacher endpoint therefore lies in the region that the closed-form flow assigns to its own data point.

\begin{table}[ht]
  \centering
  \caption{\textbf{Teacher couplings select a narrow set of endpoints within each region.} For the first $2000$ training images on each data point, the endpoints from three independent noise draws ($\lambda = 1$, $\epsilon = 10^{-3}$) are integrated with $250$ Euler steps using $\vfm$ and $\vpf$. Then, we measure the noise share, i.e. fraction of the endpoint variance explained by the injected noise, the cosine similarity between two endpoints of the same data point and fraction of endpoints whose nearest neighbour among the endpoints of another noise draw belongs to the same data point (chance level $0.05\%$).}
  \label{tab:coupling_narrowness}
  \begin{tabular}{llccc}
    \toprule
    dataset & coupling & noise share & $\cos(\x_1, \x_1')$ & NN match \\
    \midrule
    CIFAR-10 & teacher     & $6.1\%$  & $0.94$  & $100\%$ \\
             & closed-form & $99.6\%$ & $0.004$ & $0.05\%$ \\
    \midrule
    CelebA   & teacher     & $8.2\%$  & $0.92$  & $100\%$ \\
             & closed-form & $99.9\%$ & $0.001$ & $0.05\%$ \\
    \midrule
    ImageNet & teacher     & $4.3\%$  & $0.96$  & $100\%$ \\
             & closed-form & $99.8\%$ & $0.002$ & $0.03\%$ \\
    \bottomrule
  \end{tabular}
\end{table}

To compare how the two couplings select endpoints within these regions, we integrate the first $2000$ training images of each dataset from three independent noise draws through each teacher, using its best checkpoint and $250$ Euler steps, and compare them with closed-form couplings of the same images \crefp{tab:coupling_narrowness}. The two couplings behave in opposite ways. For the teacher, the injected noise explains only $4$--$8\%$ of the endpoint variance. Two endpoints of the same data point are nearly identical, with a cosine similarity of $0.92$--$0.96$, and the nearest neighbour of every teacher endpoint among the endpoints of another noise draw belongs to the same data point. For the closed-form flow, the injected noise explains more than $99\%$ of the endpoint variance. Two endpoints of the same data point are almost orthogonal, like two independent Gaussian samples, and the nearest neighbour identifies the data point only at chance level.

\section{Proofs}
\label{sec:proof}
In this section, we provide the proofs of our propositions. 

\begin{restatable}[Endpoint decomposition]{proposition}{endpointDecomposition}
\label{prop:endpoint_decomposition}
Let $(\x_t)_{t \in [\epsilon, 1]}$ solve $\ddt{\x_t} = \vpf_t(\x_t)$ with initial condition $\x_\epsilon = \alpha_\epsilon \x_0 + \lambda\, \sigma_\epsilon \z$, for arbitrary $\x_0, \z \in \mathbb{R}^d$, $\lambda \in \mathbb{R}$ and $\epsilon \in (0,1)$. Then, with $\gamma_t$ and $m_t$ as in \cref{eq:vpf_affine},
\begin{equation}
\label{eq:endpoint_identity}
\x_1 = \alpha_1 \x_0 + \lambda\, \sigma_1 \z + \sigma_1 \drift(\x_\epsilon),
\qquad
\drift(\x_\epsilon) := \int_\epsilon^1 \frac{\gamma_t}{\sigma_t}\, \big( m_t(\x_t) - \x_0 \big)\, \diff t.
\end{equation}
For the schedules of this paper ($\alpha_t = 1 - t$, $\sigma_t = t$) this reads $\x_1 = \lambda \z + \drift(\x_\epsilon)$ with $\drift(\x_\epsilon) = -\int_\epsilon^1 t^{-2} \big( m_t(\x_t) - \x_0 \big)\, \diff t$. In particular: if the posterior remains concentrated on the generating datapoint, i.e.\ $m_t(\x_t) = \x_0$ for all $t \in [\epsilon, 1]$ (e.g.\ for $N = 1$), then $\drift(\x_\epsilon) = \0$ and the flow returns the injected noise exactly, $\x_1 = \lambda \z$.
\end{restatable}
\begin{proof}
Since $\sigma_t > 0$ on $[\epsilon, 1]$ and the weights $w_t^i$ are smooth there, $\vpf$ is smooth on $[\epsilon, 1] \times \mathbb{R}^d$ and the trajectory $(\x_t)_{t \in [\epsilon,1]}$ is well defined. Recall the split of the field into its linear and nonlinear parts, \cref{eq:vpf_affine},
\begin{equation}
\vpf_t(\x) = \ddt{\log\sigma_t}\, \x + \gamma_t\, m_t(\x),
\qquad
\gamma_t = \ddt{\alpha_t} - \alpha_t \ddt{\log\sigma_t} .
\end{equation}
Define the conditional trajectory $\mathbf{y}_t := \alpha_t \x_0 + \lambda \sigma_t \z$. Direct substitution gives
\begin{equation}
\ddt{\mathbf{y}_t} = \ddt{\alpha_t}\, \x_0 + \lambda \ddt{\sigma_t}\, \z
= \ddt{\log\sigma_t}\, \mathbf{y}_t + \gamma_t\, \x_0 ,
\end{equation}
i.e.\ $\mathbf{y}_t$ solves the same ODE as $\x_t$ with $m_t(\cdot)$ replaced by the constant $\x_0$. We will derive an ODE for the deviation of the true trajectory from this reference and integrate it, this will give us the endpoint decomposition.   

Let $\mathbf{e}_t := \x_t - \mathbf{y}_t$ denote the deviation. Subtracting the two equations,
\begin{equation}
\ddt{\mathbf{e}_t} = \ddt{\log\sigma_t}\, \mathbf{e}_t + \gamma_t \big( m_t(\x_t) - \x_0 \big),
\qquad
\mathbf{e}_\epsilon = \x_\epsilon - \mathbf{y}_\epsilon = \0 .
\end{equation}
Using the integrating factor method (with the factor $\exp\big(-\int^t \frac{\diff \log\sigma_s}{\diff s}\, \diff s\big) = \frac{1}{\sigma_t}$), we get
\begin{equation}
\label{eq:relative_deviation_ode}
\ddt{\big(\frac{\mathbf{e}_t}{\sigma_t}\big)} = \frac{\gamma_t}{\sigma_t} \big( m_t(\x_t) - \x_0 \big).
\end{equation}
Integrating from $\epsilon$ to $1$ with $\mathbf{e}_\epsilon = \0$ yields
$\mathbf{e}_1 = \sigma_1 \int_\epsilon^1 \frac{\gamma_t}{\sigma_t} \big( m_t(\x_t) - \x_0 \big)\, \diff t = \sigma_1 \drift(\x_\epsilon)$.
Hence $\x_1 = \mathbf{y}_1 + \mathbf{e}_1 = \alpha_1 \x_0 + \lambda \sigma_1 \z + \sigma_1 \drift(\x_\epsilon)$.

If $m_t(\x_t) = \x_0$ for all $t \in [\epsilon, 1]$, the integrand vanishes, so $\drift(\x_\epsilon) = \0$ and $\x_1 = \lambda \z$. For $N = 1$ we have $m_t(\x) = \x_0$ identically, so this case applies with no further assumption .
\end{proof}

\begin{restatable}[Exact cutoff decomposition]{proposition}{driftCutoff}
\label{prop:drift_cutoff}
Let $\x_0 \sim \widehat p_0$ and $\z \sim \Normal(\0,\I)$ independently, and let $(\x_t)_{t\in[\epsilon,1]}$ solve the empirical probability-flow ODE from $\x_\epsilon = \alpha_\epsilon\ x_0 + \sigma_\epsilon \z$. Assume that $\alpha_t$ and $\sigma_t$ are continuously differentiable, $\sigma_t>0$ on $[\epsilon, 1]$, $\alpha_1 = 0$, $\sigma_1 = 1$, and that $u_t := \frac{\alpha_t}{\sigma_t} \geq 0$ is nonincreasing, and set $\omega_t := -\ddt{u_t} \geq 0$. Let $\bar{\x} := \frac{1}{N} \sum_{j=1}^N \x_0^j$ and $V := \frac{1}{N} \sum_{j=1}^N\|\x_0^j-\bar{\x}\|^2$ be the mean and total variance of the dataset, and write $\|Y\|_{L^2} := (\mathbb{E}_{\x_0,\z}[\|Y\|^2])^{\frac{1}{2}}$ for a random vector $Y = Y(\x_0,\z)$. For a cutoff $\tau \in (\epsilon,1)$, let $\res_\tau := \frac{1}{\sigma_\tau} \big(\x_\tau - (\alpha_\tau \x_0 + \sigma_\tau \z) \big)$ denote the normalised deviation of the trajectory from its conditional path at time $\tau$. Then, for every $\tau \in (\epsilon, 1)$, the endpoint drift from \cref{eq:endpoint_identity} satisfies the pathwise identity
\begin{equation}
\drift(\x_\epsilon) = u_\tau(\x_0 - \bar{\x}) + \res_\tau - \int_\tau^1 \omega_t\bigl(m_t(\x_t)-\bar{\x}\bigr)\,\diff t,
\end{equation}
and consequently
\begin{equation}
  \|\drift(\x_\epsilon)\|_{L^2} \leq \|\res_\tau\|_{L^2} +2u_\tau\sqrt{V}.
\end{equation}
\end{restatable}
\begin{proof}
Let $\mathbf{y}_t=\alpha_t\x_0+\sigma_t\z$ and $\mathbf{e}_t:=\x_t-\mathbf{y}_t$ be the reference path and the deviation from the proof of \cref{prop:endpoint_decomposition} with $\lambda=1$. Using
\begin{equation}
\frac{\gamma_t}{\sigma_t} = \ddt{}\left(\frac{\alpha_t}{\sigma_t}\right) = -\omega_t,
\end{equation}
we can rewrite the ODE for the relative deviation, \cref{eq:relative_deviation_ode}, as
\begin{equation}
\ddt{}\left(\frac{\mathbf{e}_t}{\sigma_t}\right) = -\omega_t\bigl(m_t(\x_t)-\x_0\bigr).
\end{equation}

At $t=1$, the reference path equals $\mathbf{y}_1=\z$, because $\alpha_1=0$ and $\sigma_1=1$. Hence $\frac{\mathbf{e}_1}{\sigma_1}=\x_1-\z=\drift(\x_\epsilon)$. At $t=\tau$, we have $\frac{\mathbf{e}_\tau}{\sigma_\tau}=\res_\tau$ by definition of $\res_\tau$. Integrating the previous identity from $\tau$ to $1$ therefore gives
\begin{equation}
\drift(\x_\epsilon) - \res_\tau = \frac{\mathbf{e}_1}{\sigma_1} - \frac{\mathbf{e}_\tau}{\sigma_\tau} = -\int_\tau^1 \omega_t\bigl(m_t(\x_t)-\x_0\bigr)\,\diff t.
\end{equation}
To isolate the dependence on $\x_0$, we write $m_t(\x_t)-\x_0 = \bigl(m_t(\x_t)-\bar{\x}\bigr) - (\x_0-\bar{\x})$ inside the integral. The second term does not depend on $t$, and the multiplier $\omega_t$ integrates to
\begin{equation}
\int_\tau^1\omega_t\,\diff t = -\int_\tau^1\ddt{u_t}\,\diff t = u_\tau-u_1 = u_\tau,
\end{equation}
since $u_1 = \frac{\alpha_1}{\sigma_1} = 0$. Solving for $\drift(\x_\epsilon)$ yields the exact decomposition
\begin{equation}
\drift(\x_\epsilon) = u_\tau(\x_0-\bar{\x}) + \res_\tau - \int_\tau^1 \omega_t\bigl(m_t(\x_t)-\bar{\x}\bigr)\,\diff t.
\end{equation}

It remains to bound the three terms in $L^2$.  Since $\x_\epsilon=\alpha_\epsilon\x_0+\sigma_\epsilon\z$, the initial state has distribution $p_\epsilon$. The probability-flow ODE therefore has marginal distribution $p_t$ at every time, so $\x_t\sim p_t$. Moreover, $m_t$ is the posterior mean under the Gaussian path.  A posterior mean cannot have more variance than the variable it estimates, and the law of total variance gives
\begin{equation}
\mathbb{E}_{X\sim p_t} \left[\|m_t(X)-\bar{\x}\|^2\right] \leq V.
\end{equation}
Applying the $L^2$ triangle inequality, also to the time integral,
\begin{equation}
\begin{aligned}
\|\drift(\x_\epsilon)\|_{L^2} & \leq u_\tau\|\x_0-\bar{\x}\|_{L^2} +\|\res_\tau\|_{L^2} +\int_\tau^1 \omega_t \left(\mathbb{E}_{X\sim p_t} \left[\|m_t(X)-\bar{\x}\|^2\right] \right)^{\frac{1}{2}} \,\diff t 
\\
&\leq u_\tau\sqrt V +\|\res_\tau\|_{L^2} +\sqrt V\int_\tau^1\omega_t\,\diff t
\\
&= \|\res_\tau\|_{L^2} +2u_\tau\sqrt V.
\end{aligned}
\end{equation}
This proves the claim.
\end{proof}

\subsection{Symmetry results}
\label{sec:proof_symmetry}

Both propositions from \cref{sec:vis} follow from one lemma, which we prove first before stating the propositions formally. If an orthogonal map permutes the data points, the closed-form field commutes with it, and the set of points it leaves fixed is invariant under the flow. Throughout, we assume that $\alpha_t$ and $\sigma_t$ are continuously differentiable.

\begin{lemma}[Equivariance]
\label{lem:equivariance}
Let $Q \in \mathbb{R}^{d \times d}$ be orthogonal and assume that $Q$ permutes the dataset, i.e.\ there is a permutation $\pi$ of $\{1, \dots, N\}$ with $Q \x_0^k = \x_0^{\pi(k)}$ for all $k$. Let $F_Q := \{\x : Q\x = \x\}$. Then
\begin{enumerate}
\item[(i)] $\vpf_t(Q\x) = Q\, \vpf_t(\x)$ for all $\x \in \mathbb{R}^d$ and $t \in (0,1]$;
\item[(ii)] $\vpf_t(\x) \in F_Q$ for all $\x \in F_Q$ and $t \in (0,1]$;
\item[(iii)] if $(\x_t)_{t \in [\epsilon, 1]}$ solves $\ddt{\x_t} = \vpf_t(\x_t)$ and $\x_s \in F_Q$ for some $s \in [\epsilon, 1]$, then $\x_t \in F_Q$ for all $t \in [\epsilon, 1]$.
\end{enumerate}
\end{lemma}
\begin{proof}
(i) Since $Q$ is orthogonal, $\|Q\x - \alpha_t \x_0^{\pi(k)}\| = \|Q\x - \alpha_t Q \x_0^k\| = \|\x - \alpha_t \x_0^k\|$, hence $\Normal(Q\x \mid \alpha_t \x_0^{\pi(k)}, \sigma_t^2 \I) = \Normal(\x \mid \alpha_t \x_0^k, \sigma_t^2 \I)$ for every $k$. Because $\pi$ is a bijection, the normalising sums in \cref{eq:vpf} agree as well, so $w_t^{\pi(k)}(Q\x) = w_t^k(\x)$. For the posterior mean of \cref{eq:vpf_affine}, this gives
\begin{equation}
m_t(Q\x) = \sum_{k=1}^N w_t^{\pi(k)}(Q\x)\, \x_0^{\pi(k)} = \sum_{k=1}^N w_t^k(\x)\, Q\x_0^k = Q\, m_t(\x),
\end{equation}
and therefore $\vpf_t(Q\x) = \ddt{\log\sigma_t}\, Q\x + \gamma_t\, Q\, m_t(\x) = Q\, \vpf_t(\x)$.

(ii) For $\x \in F_Q$, (i) gives $Q\, \vpf_t(\x) = \vpf_t(Q\x) = \vpf_t(\x)$.

(iii) Since $\vpf_t$ is continuous in $t$ and Lipschitz in $\x$ on $[\epsilon, 1]$, solutions of the \pf exist and are unique. Now let $\x_s \in F_Q$ and consider the reflected trajectory $\mathbf{y}_t := Q\x_t$. By (i),
\begin{equation}
\ddt{\mathbf{y}_t} = Q\, \vpf_t(\x_t) = \vpf_t(Q\x_t) = \vpf_t(\mathbf{y}_t),
\qquad
\mathbf{y}_s = Q\x_s = \x_s.
\end{equation}
Thus, $\mathbf{y}_t$ is itself a solution of the \pf, and at time $s$ it passes through the same point as $\x_t$, since $\x_s \in F_Q$. Two solutions that agree at one time agree at all times, so $\mathbf{y}_t = \x_t$. Hence $Q\x_t = \x_t$, i.e.\ $\x_t \in F_Q$, for all $t \in [\epsilon, 1]$.
\end{proof}

\begin{proposition}[Invariant Voronoi cells, formal version of \cref{prop:voronoi_inv}]
\label{prop:voronoi_inv_formal}
Let the data points $\{\x_0^i\}_{i=1}^N \subset \mathbb{R}^d$ be distinct with $\|\x_0^i\| = \|\x_0^j\|$ for all $i, j$, and let $V_i := \{\x : \|\x - \x_0^i\| \leq \|\x - \x_0^j\|\ \forall j\}$ be the Voronoi cell of $\x_0^i$. Call $j$ a neighbour of $i$ if $V_i$ and $V_j$ share a facet, i.e.\ $V_i \cap V_j$ has dimension $d-1$. For such a pair, let $n_{ij} := \x_0^j - \x_0^i$, let $H_{ij} := \{\x : \langle n_{ij}, \x \rangle = 0\}$ be the bisector hyperplane, and let
\begin{equation}
Q_{ij} := \I - 2 \frac{n_{ij} n_{ij}^\top}{\|n_{ij}\|^2}
\end{equation}
be the reflection across it. Assume that, for every pair of neighbours $(i,j)$, $Q_{ij}$ permutes the dataset, i.e.\ $\{Q_{ij} \x_0^k\}_{k=1}^N = \{\x_0^k\}_{k=1}^N$. Then, for every pair of neighbours, every $t \in (0,1]$ and every $\x \in H_{ij}$,
\begin{equation}
\langle n_{ij}, \vpf_t(\x) \rangle = 0,
\end{equation}
i.e.\ $\vpf_t$ is tangent to $H_{ij}$. Consequently, every Voronoi cell is invariant under the flow. If $(\x_t)_{t \in [\epsilon, 1]}$ solves $\ddt{\x_t} = \vpf_t(\x_t)$ and $\x_s \in V_i$ for some $s \in [\epsilon, 1]$, then $\x_t \in V_i$ for all $t \in [\epsilon, 1]$.
\end{proposition}
\begin{proof}
\emph{Geometry of the cells.} For any $i \neq j$,
\begin{equation}
\|\x - \x_0^i\|^2 - \|\x - \x_0^j\|^2 = 2\langle n_{ij}, \x \rangle + \|\x_0^i\|^2 - \|\x_0^j\|^2 = 2\langle n_{ij}, \x \rangle,
\end{equation}
using the equal norms. Hence $H_{ij}$ is the bisector of $\x_0^i$ and $\x_0^j$, and $V_i = \{\x : \langle n_{ij}, \x \rangle \leq 0\ \forall j \neq i\}$ is a full-dimensional convex polyhedral cone. Like every such polyhedron, it is already described by the half-spaces of its facets, $V_i = \{\x : \langle n_{ij}, \x \rangle \leq 0 \text{ for all neighbours } j \text{ of } i\}$.

\emph{Tangency.} $Q_{ij}$ is orthogonal, and its fixed-point set is $H_{ij}$. Since $Q_{ij}$ permutes the dataset, \cref{lem:equivariance}(ii) gives $\vpf_t(\x) \in H_{ij}$, i.e.\ $\langle n_{ij}, \vpf_t(\x) \rangle = 0$, for all $\x \in H_{ij}$. Moreover, by \cref{lem:equivariance}(iii), a trajectory that lies in $H_{ij}$ at one time lies in $H_{ij}$ at all times.

\emph{Invariance of the cells.} Let $\x_s \in V_i$, fix a neighbour $j$ of $i$, and consider $g(t) := \langle n_{ij}, \x_t \rangle$, which is continuous with $g(s) \leq 0$. If $g$ vanishes at some time, the trajectory lies in $H_{ij}$ at that time, and therefore at all times, so $g \equiv 0$. Otherwise, $g$ never vanishes and by continuity keeps the sign of $g(s) < 0$. In both cases, $g(t) \leq 0$ for all $t \in [\epsilon, 1]$. Since this holds for every neighbour $j$, we have $\x_t \in V_i$ for all $t \in [\epsilon, 1]$.
\end{proof}

\begin{proposition}[Invariant lines, formal version of \cref{prop:line_inv}]
\label{prop:line_inv_formal}
Fix a data point $\x_0^i \neq \0$ and let
\begin{equation}
Q_i := 2 \frac{\x_0^i (\x_0^i)^\top}{\|\x_0^i\|^2} - \I
\end{equation}
be the reflection across the line $\mathrm{span}(\x_0^i)$. Assume that $Q_i$ permutes the dataset, i.e.\ $\{Q_i \x_0^k\}_{k=1}^N = \{\x_0^k\}_{k=1}^N$. Then $\vpf_t(\x) \in \mathrm{span}(\x_0^i)$ for every $\x \in \mathrm{span}(\x_0^i)$ and $t \in (0,1]$. Consequently, the trajectory with noise-free initial condition $\x_\epsilon = \alpha_\epsilon \x_0^i$ ($\lambda = 0$) satisfies $\x_t = c_t\, \x_0^i$ for all $t \in [\epsilon, 1]$ with scalars $c_t$. In particular, the drift of the noise-free integration is an exact rescaled copy of the data point,
\begin{equation}
\drift(\alpha_\epsilon \x_0^i) = c_1\, \x_0^i.
\end{equation}
If, in addition, all data points have the same norm and the dataset has mean zero, then $c_t > 0$ for all $t \in [\epsilon, 1]$.
\end{proposition}
\begin{proof}
$Q_i$ is orthogonal, and $Q_i \x = \x$ holds if and only if $\x$ equals its projection onto $\mathrm{span}(\x_0^i)$, so its fixed-point set is $\mathrm{span}(\x_0^i)$. Since $Q_i$ permutes the dataset, \cref{lem:equivariance}(ii) gives $\vpf_t(\x) \in \mathrm{span}(\x_0^i)$ for all $\x \in \mathrm{span}(\x_0^i)$. Since $\x_\epsilon = \alpha_\epsilon \x_0^i \in \mathrm{span}(\x_0^i)$, \cref{lem:equivariance}(iii) gives $\x_t \in \mathrm{span}(\x_0^i)$, i.e.\ $\x_t = c_t\, \x_0^i$, for all $t \in [\epsilon, 1]$. Finally, \cref{prop:endpoint_decomposition} with $\lambda = 0$, $\alpha_1 = 0$ and $\sigma_1 = 1$ gives $\x_1 = \drift(\alpha_\epsilon \x_0^i)$, hence $\drift(\alpha_\epsilon \x_0^i) = c_1\, \x_0^i$.

For the sign, note that with equal norms all weights at the origin are equal, $w_t^k(\0) = 1/N$, so $m_t(\0)$ is the data mean, which is $\0$. Hence $\vpf_t(\0) = \0$, and $\x_t \equiv \0$ solves the \pf. By uniqueness, no other trajectory reaches $\0$, so $c_t$ cannot change sign, and $c_\epsilon = \alpha_\epsilon > 0$ gives $c_t > 0$ for all $t \in [\epsilon, 1]$.
\end{proof}

\section{Empirical investigation of the drift}
\label{sec:empirical_drift}

This section contains the empirical measurements to support \cref{sec:endpoint_decomposition}. \Cref{app:drift_properties} quantifies the two empirical properties of the drift on which the endpoint decomposition builds. First, the drift is small, and second, it depends only weakly on the injected noise. \Cref{app:drift_bound} evaluates both terms of the bound in \cref{prop:drift_cutoff_simple} along the trajectory, and \cref{app:drift_scaling} tests the two predictions of the scaling argument in \cref{sec:drift_scaling}. Unless stated otherwise, the empirical distribution is the full training set of each dataset, and we use $\epsilon = 10^{-3}$.

\subsection{The properties of the drift}
\label{app:drift_properties}
\begin{table}[ht]
  \centering
  \caption{\textbf{Empirical verification of \cref{eq:drift_constant}.} Every data point is integrated twice with independent noises $\z, \z' \sim \Normal(\0, \I)$ ($\lambda = 1$) and once from $\alpha_\epsilon \x_0$ (\pfmean), for $25000$ data points per dataset, with the full training set as the empirical distribution and $\epsilon = 10^{-3}$. \textbf{(a)} Share of the endpoint energy carried by the drift (first column), energy of the differences between the drifts $\drift(\x_\epsilon) = \x_1 - \z$, $\drift(\x_\epsilon') = \x_1' - \z'$ and $\drift(\alpha_\epsilon \x_0)$ relative to the drift energy $\mathbb{E}\|\drift(\x_\epsilon)\|^2$ (second and third columns), and a least-squares fit $\drift(\x_\epsilon) \approx g\,\drift(\alpha_\epsilon \x_0)$ with its gain $g$ and the energy share of the remainder (last two columns). \textbf{(b)} Cosine similarities, averaged over data points.}
  \label{tab:endpoint_decomposition}

  \textbf{(a)} Energy of the differences \\[2pt]
  \begin{tabular}{l|c|cc|cc}
    \toprule
     & & & & \multicolumn{2}{c}{fit $\drift(\x_\epsilon) \approx g\,\drift(\alpha_\epsilon \x_0)$} \\
     Dataset & $\frac{\mathbb{E}\|\drift(\x_\epsilon)\|^2}{\mathbb{E}\|\x_1\|^2}$
     & $\frac{\mathbb{E}\| \drift(\x_\epsilon) - \drift(\x_\epsilon')\|^2}{\mathbb{E}\|\drift(\x_\epsilon)\|^2}$ & $\frac{\mathbb{E}\| \drift(\x_\epsilon) - \drift(\alpha_\epsilon \x_0)\|^2}{\mathbb{E}\|\drift(\x_\epsilon)\|^2}$
     & $g$ & remainder \\
    \midrule
    CIFAR-10 & $0.52\%$ & $25.5\%$ & $15.5\%$ & $1.19$ & $13.3\%$ \\
    CelebA   & $0.14\%$ & $27.3\%$ & $16.8\%$ & $1.21$ & $14.2\%$ \\
    ImageNet & $0.18\%$ & $22.0\%$ & $14.0\%$ & $1.18$ & $12.0\%$ \\
    \bottomrule
    \label{tab:endpoint_decomposition_energy}
  \end{tabular}

  \vspace{6pt}
  \textbf{(b)} Cosine similarities \\[2pt]
  \begin{tabular}{l|cc|cc}
    \toprule
     Dataset & $\cos(\drift(\x_\epsilon), \drift(\x_\epsilon'))$ & $\cos(\drift(\x_\epsilon), \drift(\alpha_\epsilon \x_0))$
     & $\cos(\drift(\x_\epsilon), \x_0)$ & $\cos(\drift(\alpha_\epsilon \x_0), \x_0)$ \\
    \midrule
    CIFAR-10 & $0.899$ & $0.945$ & $0.867$ & $0.932$ \\
    CelebA   & $0.888$ & $0.939$ & $0.746$ & $0.796$ \\
    ImageNet & $0.911$ & $0.952$ & $0.864$ & $0.923$ \\
    \bottomrule
    \label{tab:endpoint_decomposition_cosine}
  \end{tabular}
\end{table}

\Cref{tab:endpoint_decomposition} is based on three integrations per data point. Two start from independent noisy initial conditions ($\lambda = 1$) and give the drifts $\drift(\x_\epsilon) = \x_1 - \z$ and $\drift(\x_\epsilon') = \x_1' - \z'$. The third starts from the noise-free initial condition $\alpha_\epsilon \x_0$ and gives the \pfmean endpoint $\drift(\alpha_\epsilon \x_0)$.

\paragraph{The drift is small.} On all three datasets, the drift carries well below $1\%$ of the endpoint energy \crefp{tab:endpoint_decomposition}(a). In absolute terms, its norm is nearly the same across datasets, between $4.0$ and $4.7$ \crefp{fig:drift_bound_evaluation}, although the dimension of CelebA and ImageNet-64 is four times that of CIFAR-10. The smaller share on these two datasets therefore reflects the larger dimension rather than a smaller drift. This is the $\frac{1}{d}$ dependence that \cref{sec:drift_scaling} predicts and \cref{sec:latent} exploits.

\paragraph{The drift depends only weakly on the noise.} The least-squares fit in \cref{tab:endpoint_decomposition}(a) shows that, on every dataset, the drift is an amplified copy of the noise-free drift, with gain $g \approx 1.2$, and the remainder carries only $12$--$14\%$ of the drift energy. The cosine similarities in \cref{tab:endpoint_decomposition}(b) show that this remainder is the part that changes with the noise. Each noisy drift is closely aligned with the noise-free drift, with $\cos(\drift(\x_\epsilon), \drift(\alpha_\epsilon \x_0)) = 0.94$--$0.95$. If every noisy drift consists of the noise-free drift plus an independent noise-specific part, $\drift(\x_\epsilon)$ and $\drift(\x_\epsilon')$ would share only the noise-free drift, and their cosine similarity should be approximately the square of this value, $0.88$--$0.91$. The measured $\cos(\drift(\x_\epsilon), \drift(\x_\epsilon'))$ of $0.89$--$0.91$ matches this closely. Therefore, $\drift(\alpha_\epsilon \x_0)$ captures what the noisy drifts have in common, and only the small remainder varies with the noise. This supports the approximation in \cref{eq:drift_constant}, up to the gain $g$.

\paragraph{The drift carries the data point.} Finally, both drifts are well aligned with the data point itself (\cref{tab:endpoint_decomposition}(b)). The noise-free drift is consistently the better aligned of the two, as expected if the noise-specific remainder carries little information about $\x_0$. The alignment is lowest on CelebA, but even there the drift retains much of the structure of its data point. \Cref{fig:drift_constant_visualisation_single,fig:drift_constant_visualisation} show what these numbers look like on ImageNet-64.

\subsection{The bound on the drift investigated}
\label{app:drift_bound}
\begin{figure}[ht]
  \centering
  \includegraphics[width=\linewidth]{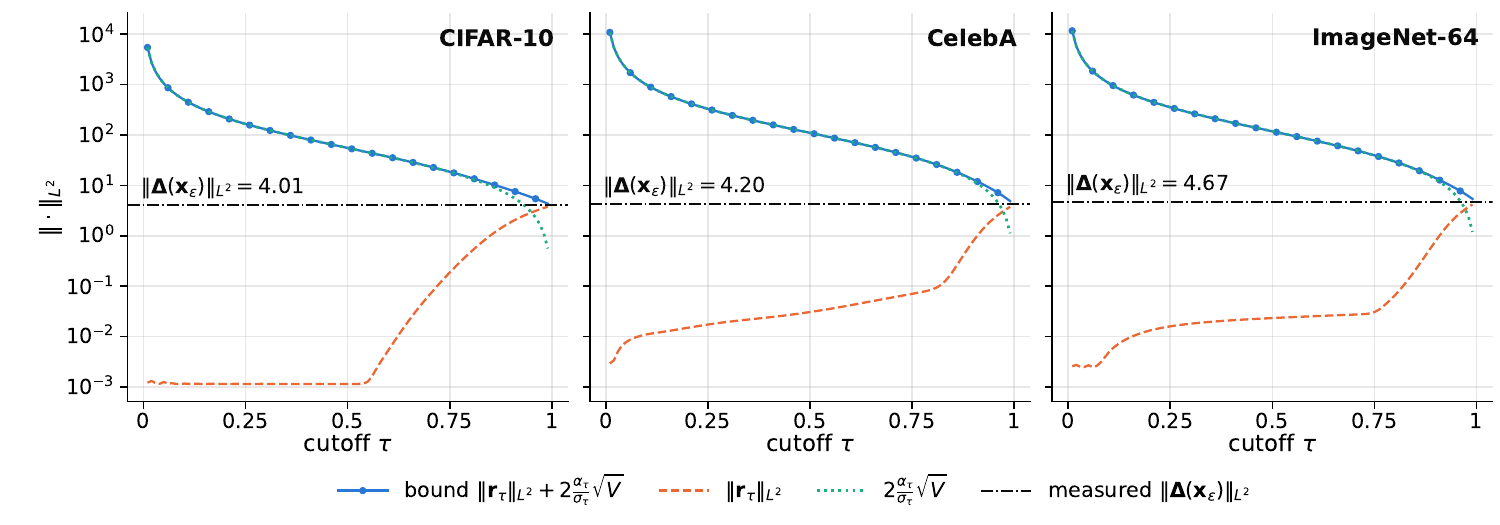}
  \caption{\textbf{Evaluation of the drift bound.} We evaluate both terms of \cref{prop:drift_cutoff_simple} on all three datasets for different values of the cutoff $\tau$, using the entire training set. The bound becomes increasingly tight as $\tau$ increases, showing that \cref{prop:drift_cutoff_simple} provides a practical bound on the drift of the \pf trajectory. Furthermore, the tracking error $\|\res_\tau\|_{L^2}$ remains negligible for small $\tau$, supporting the assumption in \cref{eq:delta_approx}.}
  \label{fig:drift_bound_evaluation}
\end{figure}

\Cref{fig:drift_bound_evaluation} evaluates the two terms of \cref{prop:drift_cutoff_simple} as a function of the cutoff $\tau$, together with the measured drift. 

The two terms trade off along the trajectory. For small $\tau$, the trajectory tracks the conditional path of its data point almost perfectly. The tracking error $\|\res_\tau\|_{L^2}$ stays between $10^{-3}$ and $10^{-1}$, and the bound is dominated by the SNR term $2\frac{\alpha_\tau}{\sigma_\tau}\sqrt{V}$, which is still large because the SNR is high. Only late in the trajectory, from $\tau \approx 0.55$ on CIFAR-10, $\tau \approx 0.8$ on CelebA and $\tau \approx 0.75$ on ImageNet-64, does the tracking error rise sharply as the trajectory leaves the conditional path. By then, the SNR term has decayed, and the bound approaches the measured drift. At $\tau = 1$ the bound is exact, since $\alpha_1 = 0$ and $\res_1 = \x_1 - \z = \drift(\x_\epsilon)$.

This supports the step from \cref{prop:drift_cutoff_simple} to \cref{eq:delta_approx} in \cref{sec:drift_scaling}. Before the trajectory stops tracking its data point, the tracking error is negligible, and the drift is controlled by the SNR term alone. At such $\tau$, the bound exceeds the measured drift by roughly an order of magnitude, so it captures how the drift scales rather than its exact value.

\subsection{Scaling of the drift contribution}
\label{app:drift_scaling}
\begin{figure}
\centering
\begin{subfigure}{.5\textwidth}
  \centering
  \includegraphics[width=\linewidth]{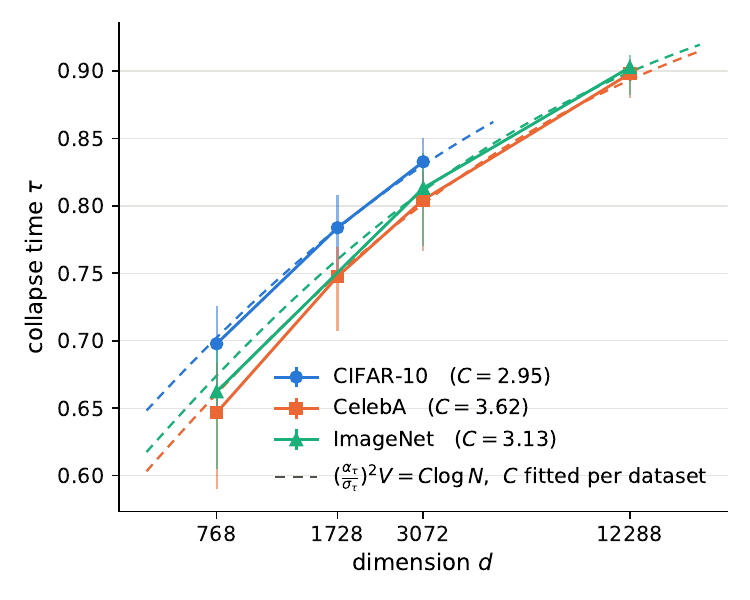}
  \label{subfig:posterior_collapse}
\end{subfigure}%
\begin{subfigure}{.5\textwidth}
  \centering
  \includegraphics[width=\linewidth]{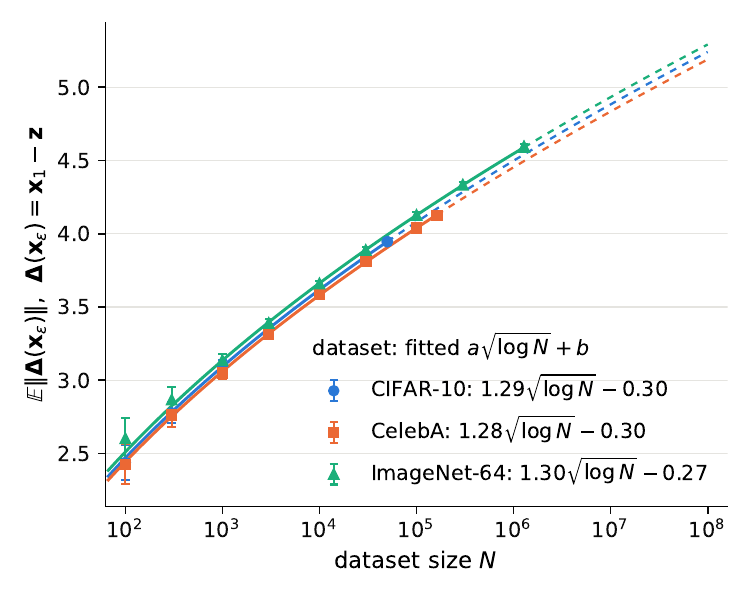}
  \label{subfig:drift_vs_N}
\end{subfigure}
\caption{\textbf{Empirical validation of the scaling argument.} \textbf{Left:} The posterior collapse time $\tau$ as the data dimension is varied by downsampling. Dashed lines show the prediction $(\frac{\alpha_\tau}{\sigma_\tau})^2V = C \log N$ from \cref{eq:posterior_collapse}, with a single constant $C$ fitted per dataset, which closely tracks the measured transition across CIFAR-10, CelebA and ImageNet-64. \textbf{Right:} Drift magnitude as a function of the dataset size $N$. Solid lines show fits of the form $\mathbb{E}\|\drift(\x_\epsilon)\| = a \sqrt{\log N} + b$, confirming the $\sqrt{\log N}$ scaling predicted by \cref{eq:delta_approx}. Dashed lines extrapolate the fits beyond the full training set, and error bars are 95\% confidence intervals.}
\label{fig:emp_val}
\end{figure}

\Cref{fig:emp_val} tests the two predictions of \cref{sec:drift_scaling}, specifically the collapse criterion in \cref{eq:posterior_collapse} and the resulting $\sqrt{\log N}$ growth of the drift in \cref{eq:delta_approx}.

\paragraph{Collapse time.} To measure the collapse time, we follow the conditional path $\x_t = \alpha_t \x_0^i + \sigma_t \z$ for $10000$ data points and evaluate the posterior weights $w_t^j(\x_t)$ of \cref{eq:vpf} against the full training set. The collapse time $\tau$ is the first $t$ at which the mean maximum weight $\max_j w_t^j(\x_t)$ falls below $\frac{1}{2}$. For the prediction, we solve \cref{eq:posterior_collapse} for $\tau$, using the measured data variance $V$ at each resolution and fitting a single constant $C$ per dataset. For our linear schedule, this gives $\tau = \big(1 + \sqrt{C \log N / V}\big)^{-1}$. Lowering the dimension reduces $V$ and with it the distance penalty that keeps the posterior on its data point, so the posterior collapses earlier. The collapse time drops from $\tau \approx 0.9$ at $d = 12288$ to $\tau \approx 0.65$--$0.7$ at $d = 768$, and the prediction follows this trend on every dataset with $C$ between $2.95$ and $3.62$ (\cref{fig:emp_val}, left).

\paragraph{Drift magnitude.} To vary $N$, we restrict the empirical distribution to random subsets of the training set, from $N = 10^2$ up to the full dataset. Over four orders of magnitude in $N$, the mean drift norm grows only from about $2.5$ to $4.6$ and is well described by $a\sqrt{\log N} + b$ with $a \approx 1.3$ on all three datasets (\cref{fig:emp_val}, right). The three curves nearly coincide, even though the datasets differ in content and CIFAR-10 has a quarter of the dimension of the other two. This is consistent with \cref{eq:delta_approx}, in which the drift depends on $N$ but not on $d$, and with the nearly constant drift norm across resolutions in \cref{fig:dimension_ladder}. Extrapolating the fits shows how little more data would help. Even at $N = 10^8$, the drift norm would reach only about $5.3$, which at $d = 12288$ is still only about $0.2\%$ of the endpoint energy.

\section{Experimental Details}
\label{sec:exp_detail}
This section describes our experimental setup. We first describe the datasets and how we obtain the different resolutions, then the flow-matching teachers, the construction of the three coupling sets, and finally the training and evaluation of the rectified-flow students.

\subsection{Datasets and resolutions}
\label{app:data_and_res}
We use the training sets of CIFAR-10 \citep{krizhevsky2009learning}, CelebA \citep{liu2015deep} (the standard training partition of the aligned-and-cropped release) and ImageNet-1k \citep{russakovsky2015imagenet}, all unconditional, without augmentation, with pixel values scaled to $[-1, 1]$. The dimension $d = 3\,r^2$ is varied by changing the resolution $r$. All resolutions of a dataset are generated from the same source images, so neighbouring resolutions differ in nothing but resolution.

\paragraph{Downsampling.} For CelebA, we take the aligned $178 \times 218$ images, crop the central $140 \times 140$ pixels and resize them with bilinear interpolation to $16$, $32$ and $64$ px. ImageNet images are centre-cropped to a square and box-downsampled to $64$ px, following \cite{chrabaszcz2017downsampled}. The $32$ and $16$ px versions are $2 \times 2$ and $4 \times 4$ box averages of these $64$ px images. CIFAR-10 is only available at $32$ px. We obtain the $16$ px version as a $2 \times 2$ box average, and the $24$ px version, a non-integer factor of $4/3$, with PIL's area filter, which generalises the box average to non-integer factors. Across all resolutions, the per-pixel standard deviation stays between $0.48$ and $0.55$, so we use the same preconditioning constant $\sigma_{\mathrm{data}} = 0.5$ throughout.

\subsection{Flow-matching teachers}
\label{app:details_teacher}
All teachers are trained with the flow-matching objective on the linear path $\alpha_t = 1 - t$, $\sigma_t = t$, with the EDM preconditioning of \cite{karras2022elucidating} around the network ($\sigma_{\mathrm{data}} = 0.5$), the EDM loss weight, and timesteps $t$ drawn from a logit-normal distribution in log-SNR (mean $2.5$, standard deviation $2.5$). Optimisation uses Adam, an exponential moving average of the weights with rate $0.9999$, and a linear learning-rate warm-up over the first $20000$ steps. Model selection uses the checkpoint with the lowest FID-50k at 32 Euler steps. Teacher quality (FID-50k, best checkpoint) is reported in \cref{tab:teacher_fid} at 64 and 128 Euler steps. To have a fair comparison between the teacher model and the \pf, we train the model without data augmentation.

\paragraph{CIFAR-10 and CelebA.} The network is the DDPM++ UNet of \cite{songscore} following the exact configuration used in \cite{karras2022elucidating} for CIFAR-10. At $16$, $24$ and $32$ px, this network is used unchanged (only the spatial size of its feature maps changes). However, at $64$ px for CelebA we use EDM's FFHQ-64 variant with channel multipliers $[1, 2, 2, 2]$ and attention at $16 \times 16$. Teachers train for $300000$ steps at global batch size $512$ with learning rate $10^{-3}$ ($2 \times 10^{-4}$ with gradient clipping at norm $1$ for the $64$ px teacher, following EDM's 64-px setting).

\paragraph{ImageNet.} The network is the ADM UNet of \cite{dhariwal2021diffusion} following \cite{karras2022elucidating} ImageNet-64 configuration. The identical network is used at $16$, $32$ and $64$ px, so the three ImageNet rungs share their capacity exactly. Only the attention resolutions scale with the image. Teachers train with learning rate $10^{-4}$ at global batch size $1024$ in bf16 mixed precision for $400000$ steps ($600000$ at $64$ px).

\subsection{Couplings}
\label{app:details_couplings}
Every coupling set pairs each of the $N$ training images $\x_0^i$ with one noise endpoint $\x_1^i$ and the students are then trained on these fixed pairs only.

\paragraph{Random couplings.} To construct $\crnd$ we draw $\x_1^i \sim \Normal(\0, \I)$ and couple them independently with $\x_0^i$. 

\paragraph{Closed-form couplings.} For $\cpf$ we start from $\x_\epsilon^i \sim \Normal(\alpha_\epsilon \x_0^i, \sigma_\epsilon^2 \I)$ with $\epsilon = 10^{-3}$ and $\lambda = 1$ and integrate the closed-form probability-flow field of the empirical distribution of all $N$ training images from $t = \epsilon$ to $t = 1$ with an adaptive Dormand–Prince solver (relative tolerance $10^{-8}$, absolute tolerance $10^{-10}$). The field in \cref{eq:vpf} is evaluated exactly against all $N$ training images, without the nearest-neighbour approximations used in \cite{scarvelis2025closedform,bertrand2025closed}. The injected noise $\z$ is then used to measure the drift $\drift = \x_1 - \z$. Each field evaluation touches the whole training set, which makes it memory intensive. Therefore, for CelebA and ImageNet the $N$ trajectories are integrated in chunks against the full set of $N$ guiding points held on the GPU.

\paragraph{Teacher couplings.} To obtain $\cfmf$ we integrate the same initial states
$\x_\epsilon^i$ forward through the flow-matching teacher's EMA weights with Euler steps on a uniform time grid to $t = 1$. For each dataset we use $128$ Euler steps from $\epsilon = 10^{-3}$ and the teacher's best checkpoint.

\subsection{Rectified-flow students}
\label{app:details_students}
For every coupling set one rectified-flow student is trained on the fixed pairs
$(\x_0^i, \x_1^i)$ with the flow-matching loss along the straight interpolant. The timesteps $t$ are drawn from the tilted-uniform distribution of \cite{lee2024improving} with tilt $a = 4$ (antithetic, symmetric). We use Adam with $5000$ warm-up steps and an EMA of rate $0.9999$. The student has the same architecture as the teacher of its resolution. On CIFAR-10 and CelebA the students use learning rate $2 \times 10^{-4}$, global batch $512$ and the EDM loss weight. The students on random and closed-form couplings train for $200000$ steps and students on teacher couplings for $100000$ steps, after which their FID no longer improves. On ImageNet all students use learning rate $10^{-4}$, global batch $512$, no loss weighting, bf16 mixed precision and $200000$ steps. Students never see augmented data, since the coupling fixes the pairing.

\paragraph{Repeats.}
To estimate the robustness of our findings, we repeat experiments on CIFAR-10 and CelebA for three seeds. Each repeat $k \in \{1, 2, 3\}$ draws fresh random couplings and fresh injected noise for the closed-form couplings with seed $k$ and trains its students with seed $41 + k$. The teacher couplings are the same file in all three repeats, so their repeats differ only in the student's initialisation and data order.

\paragraph{Evaluation.}

All FIDs are estimated using $50000$ samples against the Inception-v3 pool-3 statistics of the full training set at the same resolution. Students are sampled with the Euler method on a uniform time grid with $k \in \{1, 2, 4, 8, 16, 32\}$ steps from $t = 1$ to $t = 10^{-3}$, using the EMA weights, and every saved checkpoint (every $25000$ steps) is evaluated separately at each $k$. Teachers are sampled the same way with $64$ and $128$ Euler steps.

The gap score $G$ in \cref{eq:relative_score} compares the three students of a setting at equal $k$. $G = 0$ means the closed-form student is no better than the random one, and $G = 1$ that it matches the student trained on the teacher's couplings. Since the three students of a setting share the architecture, optimisation settings and data, $G$ is comparable across the rungs of one dataset, however across datasets it is not, because the score depends heavily on the random baseline. For further discussion see \cref{app:pair_fitting}.

\section{Additional results on reducing the dimension}
\label{app:additional_figures}

In this section we provide the results from \cref{sec:latent} in more detail. We first confirm that the teachers used to generate $\cfmf$ are strong, and then report the FIDs of the three students, from which the gap score is computed. We then show that lowering the dimension makes closed-form couplings more informative, both in how closely each endpoint aligns with its data point and in how much of the data point can be recovered from it, whereas teacher couplings are highly informative at every resolution. Finally, we show that students trained on closed-form couplings use this information beyond their training pairs.

\begin{table}[ht]
  \centering
  \caption{\textbf{FID of the flow-matching teachers used to generate $\cfmf$.} FID of the best checkpoint at $64$ and $128$ Euler steps.}
  \label{tab:teacher_fid}
    \begin{tabular}{llrrr}
    \toprule
    dataset & px & $d$ & 64 steps & 128 steps \\
    \midrule
    CIFAR-10 & 16 & 768 & 2.6 & 2.1 \\
    CIFAR-10 & 24 & 1728 & 3.4 & 3.1 \\
    CIFAR-10 & 32 & 3072 & 3.6 & 3.1 \\
    \midrule
    CelebA & 16 & 768 & 1.5 & 1.2 \\
    CelebA & 32 & 3072 & 2.6 & 2.7 \\
    CelebA & 64 & 12288 & 4.4 & 3.4 \\
    \midrule
    ImageNet & 16 & 768 & 1.8 & 1.3 \\
    ImageNet & 32 & 3072 & 3.8 & 3.2 \\
    ImageNet & 64 & 12288 & 8.1 & 7.2 \\
    \bottomrule
    \end{tabular}
\end{table}

\paragraph{The teachers are strong.} All teachers reach low FIDs \crefp{tab:teacher_fid}, between $1.2$ and $3.4$ at $128$ Euler steps in every setting except ImageNet-64, the hardest one, where they reach $7.2$. These numbers are competitive given that we have not used data augmentation or class-conditional generation.

\begin{table}[ht]
  \centering
  \caption{\textbf{FID of the students presented in \cref{fig:dimension_ladder}.} One- and two-step FID of the students trained on $\crnd$, $\cpf$ and $\cfmf$, from which the gap score $G$ \crefp{eq:relative_score} is computed. For CIFAR-10 and CelebA we report mean $\pm$ standard deviation over three seeds.}
  \label{tab:latent_fid}
    \begin{tabular}{llrrrrrr}
    \toprule
     &  &  \multicolumn{2}{c}{$\crnd$} & \multicolumn{2}{c}{$\cpf$} & \multicolumn{2}{c}{$\cfmf$} \\
    \cmidrule(lr){3-4} \cmidrule(lr){5-6} \cmidrule(lr){7-8}
    dataset & px & 1 step & 2 steps & 1 step & 2 steps & 1 step & 2 steps \\
    \midrule
    CIFAR-10 & 16 & 61.0 $\pm$ 1.9 & 38.1 $\pm$ 0.7 & 31.3 $\pm$ 1.2 & 20.2 $\pm$ 1.1 & 4.6 $\pm$ 0.0 & 3.9 $\pm$ 0.0\\
    CIFAR-10 & 24 & 72.7 $\pm$ 0.9 & 49.1 $\pm$ 0.6 & 51.5 $\pm$ 1.5 & 36.5 $\pm$ 1.1 & 9.3 $\pm$ 0.2 & 8.3 $\pm$ 0.3\\
    CIFAR-10 & 32 & 82.9 $\pm$ 0.7 & 60.7 $\pm$ 6.1 & 68.0 $\pm$ 2.3 & 46.5 $\pm$ 2.5 & 15.2 $\pm$ 0.5 & 12.4 $\pm$ 0.3\\
    \midrule
    CelebA & 16 & 38.3 $\pm$ 1.0 & 29.2 $\pm$ 0.8 & 21.3 $\pm$ 0.1 & 14.7 $\pm$ 0.1 & 2.0 $\pm$ 0.1 & 2.0 $\pm$ 0.0\\
    CelebA & 32 & 43.1 $\pm$ 2.1 & 29.6 $\pm$ 1.1 & 29.1 $\pm$ 1.8 & 20.2 $\pm$ 0.7 & 4.8 $\pm$ 0.0 & 6.4 $\pm$ 0.2\\
    CelebA & 64 & 70.4 $\pm$ 0.4 & 50.7 $\pm$ 0.4 & 53.1 $\pm$ 0.2 & 38.4 $\pm$ 0.0 & 12.3 $\pm$ 0.6 & 8.3 $\pm$ 0.3 \\
    \midrule
    ImageNet & 16 & 430.1 & 158.0 & 97.5 & 44.2 & 2.3 & 1.6\\
    ImageNet & 32 & 417.8 & 184.1 & 145.1 & 59.0 & 9.1 & 6.3\\
    ImageNet & 64 & 415.5 & 175.6 & 164.6 & 81.2 & 29.7 & 22.3 \\
    \bottomrule
    \end{tabular}
\end{table}

\paragraph{FIDs behind the gap score.} \Cref{tab:latent_fid} shows the one- and two-step FIDs from which the gap score in \cref{fig:dimension_ladder} is computed. The ordering is the same in every resolution. In each setting, the $\cfmf$ student is best, the $\crnd$ student worst, and the $\cpf$ student in between. Absolute FIDs, however, are not comparable across resolutions, which is why \cref{sec:latent} compares the students through $G$. At the lowest resolution, the $\cfmf$ students come close to their teachers, with one-step FIDs of $2.0$--$4.6$ against $1.2$--$2.1$ for the teachers at $128$ steps. On ImageNet, the random students reach FIDs above $400$ at one step. They cannot fit their $1.28$M random training pairs and simply predict the dataset mean, which inflates $G$ on this \crefp{app:pair_fitting}.

\begin{table}[ht]
  \centering
  \caption{\textbf{Alignment between data points and their coupled endpoints.} Cosine similarity $\cos(\x_0, \x_1)$ for the three couplings, averaged over data points. The last column is the alignment of the closed-form drift with its data point, $\cos(\drift(\x_\epsilon), \x_0)$.}
  \label{tab:cosine}
    \begin{tabular}{llrrrrc}
    \toprule
    dataset & px & $d$ & $\crnd$ & $\cpf$ & $\cfmf$ & $\cos(\drift(\x_\epsilon), \mathbf{x}_0)$ \\
    \midrule
    CIFAR-10 & 16 & 768 & 0.000 & 0.120 & 0.335 & 0.852 \\
    CIFAR-10 & 24 & 1728 & 0.000 & 0.082 & 0.284 & 0.861 \\
    CIFAR-10 & 32 & 3072 & 0.000 & 0.062 & 0.230 & 0.867 \\
    \midrule
    CelebA & 16 & 768 & 0.000 & 0.103 & 0.242 & 0.703 \\
    CelebA & 32 & 3072 & 0.000 & 0.054 & 0.171 & 0.730 \\
    CelebA & 64 & 12288 & 0.000 & 0.028 & 0.108 & 0.746 \\
    \midrule
    ImageNet & 16 & 768 & 0.000 & 0.135 & 0.360 & 0.826 \\
    ImageNet & 32 & 3072 & 0.000 & 0.070 & 0.271 & 0.850 \\
    ImageNet & 64 & 12288 & 0.000 & 0.036 & 0.204 & 0.864 \\
    \bottomrule
    \end{tabular}
\end{table}

\paragraph{Lower dimension makes closed-form couplings more informative.} \Cref{tab:cosine} measures how closely each endpoint aligns with its data point. For random couplings, the cosine similarity is zero, as expected for independent high-dimensional vectors. For closed-form couplings, it is small but grows as the dimension decreases, almost fourfold from $64$ to $16$ px on CelebA and ImageNet and twofold from $32$ to $16$ px on CIFAR-10, in line with the $\frac{1}{\sqrt{d}}$ scaling derived in \cref{sec:latent}. The last column shows why. The alignment of the drift with its data point stays roughly constant across resolutions, so the drift carries similar information at every resolution, and only its weight relative to the injected noise changes. Teacher couplings, in contrast, are strongly aligned at every resolution, roughly four to six times more than closed-form couplings at the native resolution. Even at $16$ px, closed-form couplings remain less aligned than teacher couplings at the highest resolution, except on CelebA, where the two are about equal.

\begin{table}[ht]
  \centering
  \caption{\textbf{Linear read-out of the data point from the coupled endpoint.} Held-out and training $R^2$ of a ridge regression $\x_0 - \bar{\x} \approx \mathbf{W}^\top \x_1$, i.e.\ the fraction of the data variance a linear map recovers from the endpoint $\x_1$ (0: predicts the mean image, 1: exact recovery). Fitted on the first $40000$ (CIFAR-10) or $100000$ (CelebA, ImageNet) pairs, tested on the next $10000$.}
  \label{tab:linear_readout}
  \begin{tabular}{llr rr rr rr}
    \toprule
     &  &  & \multicolumn{2}{c}{$\crnd$} & \multicolumn{2}{c}{$\cpf$} & \multicolumn{2}{c}{$\cfmf$} \\
    \cmidrule(lr){4-5} \cmidrule(lr){6-7} \cmidrule(lr){8-9}
    dataset & px & $d$ & train & test & train & test & train & test \\
    \midrule
    CIFAR-10 & 16 & 768 & 0.01 & 0.00 & 0.54 & 0.52 & 0.86 & 0.85 \\
    CIFAR-10 & 24 & 1728 & 0.03 & -0.01 & 0.52 & 0.48 & 0.84 & 0.82 \\
    CIFAR-10 & 32 & 3072 & 0.05 & -0.01 & 0.52 & 0.45 & 0.81 & 0.78 \\
    \midrule
    CelebA & 16 & 768 & 0.00 & 0.00 & 0.60 & 0.60 & 0.84 & 0.84 \\
    CelebA & 32 & 3072 & 0.02 & -0.01 & 0.55 & 0.53 & 0.79 & 0.78 \\
    CelebA & 64 & 12288 & 0.09 & -0.03 & 0.56 & 0.45 & 0.80 & 0.74 \\
    \midrule
    ImageNet & 16 & 768 & 0.00 & 0.00 & 0.56 & 0.56 & 0.85 & 0.85 \\
    ImageNet & 32 & 3072 & 0.02 & 0.00 & 0.51 & 0.48 & 0.82 & 0.81 \\
    ImageNet & 64 & 12288 & 0.09 & -0.02 & 0.52 & 0.39 & 0.81 & 0.75 \\
    \bottomrule
  \end{tabular}
\end{table}

\paragraph{The information becomes easier to extract.} \Cref{tab:linear_readout} asks how much of the data point a linear map can recover from the endpoint. Random endpoints carry no information, and the held-out $R^2$ is zero. For closed-form couplings, the held-out $R^2$ rises as the dimension decreases, from $0.45$ to $0.52$ on CIFAR-10, from $0.45$ to $0.60$ on CelebA and from $0.39$ to $0.56$ on ImageNet. Since the drift keeps its alignment with the data point, this gain comes from the noise masking less of it. Teacher couplings reach $0.74$--$0.85$ at every resolution, above the best closed-form value in any setting. At the highest resolutions, the training $R^2$ noticeably exceeds the held-out one, e.g.\ $0.52$ against $0.39$ for closed-form couplings on ImageNet-64. The ridge regression starts to overfit, as the non-zero training $R^2$ of random couplings in the same setting shows, so the held-out values are the relevant ones.

\begin{table}[ht]
  \centering
  \caption{\textbf{Reconstruction of held-out images.} $R^2 = 1 - \mathbb{E}\|\hat{\x}_0 - \x_0\|^2 / V$ of the one-step student output $\hat{\x}_0$ against the paired image, on the student's own training couplings (train) and on couplings built the same way for held-out images (test). $V$ is the variance of the training images, so $R^2 = 0$ is the accuracy of predicting the mean image. Test couplings of $\cpf$ integrate the closed-form field of the training images together with the held-out images.}
  \label{tab:test_reconstruction}
  \begin{tabular}{llrrrrrrr}
    \toprule
     &  &  & \multicolumn{2}{c}{$\crnd$} & \multicolumn{2}{c}{$\cpf$} & \multicolumn{2}{c}{$\cfmf$} \\
    \cmidrule(lr){4-5} \cmidrule(lr){6-7} \cmidrule(lr){8-9}
    dataset & px & $d$ & train & test & train & test & train & test \\
    \midrule
    CIFAR-10 & 16 & 768 & 0.96 & $-$0.54 & 0.97 & 0.44 & 1.00 & 0.98 \\
    CIFAR-10 & 24 & 1728 & 0.95 & $-$0.51 & 0.96 & 0.40 & 1.00 & 0.96 \\
    CIFAR-10 & 32 & 3072 & 0.92 & $-$0.59 & 0.92 & 0.33 & 0.99 & 0.92 \\
    \midrule
    CelebA & 16 & 768 & 0.95 & $-$0.52 & 0.97 & 0.61 & 1.00 & 0.99 \\
    CelebA & 32 & 3072 & 0.89 & $-$0.50 & 0.92 & 0.52 & 1.00 & 0.98 \\
    CelebA & 64 & 12288 & 0.85 & $-$0.47 & 0.92 & 0.47 & 0.99 & 0.98 \\
    \midrule
    ImageNet & 16 & 768 & 0.62 & $-$0.35 & 0.82 & 0.51 & 1.00 & 0.99 \\
    ImageNet & 32 & 3072 & 0.17 & $-$0.06 & 0.78 & 0.43 & 0.99 & 0.98 \\
    ImageNet & 64 & 12288 & 0.02 & $-$0.01 & 0.65 & 0.40 & 0.99 & 0.97 \\
    \bottomrule
  \end{tabular}
\end{table}

\paragraph{Students use the drift beyond their training pairs.} Finally, \cref{tab:test_reconstruction} tests whether the students exploit this information, or merely memorise their training pairs. We build couplings for held-out images in the same way and compare the one-step output of each student with the paired image. Students trained on closed-form couplings reconstruct $33$--$61\%$ of the held-out variance, and again more at lower dimensions. This is similar to what the linear read-out recovers from the same kind of endpoints, whereas students trained on teacher couplings reconstruct at least $92\%$ at every resolution. Random students behave differently. On CIFAR-10 and CelebA, they reproduce their training pairs with $R^2 \geq 0.85$, but reach a negative $R^2$ on held-out images, because they output plausible images that are unrelated to the paired data point. On ImageNet, they fit their training pairs only partially at $16$ px and hardly at all at $64$ px, which we discuss in \cref{app:pair_fitting}.

\section{Drift energy and learnability}
\label{app:learnability}

The endpoint decomposition gives us a simple picture. The closed-form flow returns the injected noise plus a drift that, in our experiments, is small, resembles its originating image, and depends only weakly on the injected noise. This small drift can contain much information about the image, but it might be difficult to recover this structure from the endpoint, where the drift is mixed with noise. In the following, we examine how the drift share in \cref{eq:drift_share} relates to the accessibility of this structure, and clarify its usefulness and limitations as a proxy for coupling learnability. 
As a simple measure of learnability, we use the fraction of image variance recovered by the best linear predictor from the endpoint. This does not characterise what every nonlinear model can learn, but on \cref{tab:linear_readout,tab:test_reconstruction} we saw that our empirical linear readouts follow the reconstruction values achieved by the closed-form students.

Let $\mathbf{u}_k$ be the principal directions of the data, with variances $\lambda_k$, total variance $V=\sum_k\lambda_k>0$, and variance shares $\mu_k=\lambda_k/V$. Write $s_k=\mathbf{u}_k^\top(\x_0-\bar{\x})$ for the $k$-th data component, and $e_k = \mathbf{u}_k^\top \x_1$ and $\delta_k = \mathbf{u}_k^\top \drift(\x_\epsilon)$ for the endpoint and drift components in the same basis. Since $\x_1\sim\Normal(\0,\I)$, the endpoint components are uncorrelated and have unit variance. For each $\lambda_k>0$, let
\begin{equation}
    r_k = \sum_{j=1}^d \frac{\operatorname{Cov}(s_k, e_j)^2}{\lambda_k} = \sum_{j=1}^d \frac{\operatorname{Cov}(s_k, \delta_j)^2}{\lambda_k}\in[0,1]
\end{equation}
be the fraction of its variance recovered by the best linear predictor from the full endpoint $\x_1$. The second equality holds because the injected noise $\z$ is independent of $\x_0$, so only the drift contributes to the covariance. Thus, $r_k=1$ means that the component can be recovered perfectly, while $r_k=0$ means that a linear predictor does no better than predicting its mean.

The optimal linear reconstruction recovers
\begin{equation}
  \label{eq:readout}
  R^2_{\mathrm{lin}} = \sum_k \mu_k r_k = \frac{\|\operatorname{Cov}(\x_0, \drift(\x_\epsilon))\|_F^2}{V}
\end{equation}
of the total data variance, where the second equality follows from the invariance of the Frobenius norm under the orthogonal change of basis. Throughout, sums over data components include only $\lambda_k>0$; components with zero variance are constant, are already reconstructed by the mean, and do not contribute to the covariance.

Since the injected noise is independent of the data, all linear dependence between the endpoint and the image comes from the drift. This gives the following connection.

\begin{proposition}[Drift energy and linear recovery]
  \label{prop:readable_directions}
  Let $\x_1=\z+\drift(\x_\epsilon)$ be the endpoint of the closed-form flow from \cref{prop:endpoint_decomposition}. Let $\widehat{\drift}$ be the best linear predictor of $\drift(\x_\epsilon)$ from $\x_0-\bar{\x}$, and define
  \begin{equation}
    \label{eq:eta}
    \eta :=
    \frac{\mathbb{E}\|\widehat{\drift}\|^2}{\energy{\drift(\x_\epsilon)}} \in [0,1]
  \end{equation}
  as the fraction of drift energy linearly explained by the data, with $\eta=0$ if the drift energy is zero. Then
  \begin{equation}
    \label{eq:readable_directions}
    \sum_k r_k = \eta\,\energy{\drift(\x_\epsilon)} \le \energy{\drift(\x_\epsilon)}.
  \end{equation}
\end{proposition}

\begin{proof}
  Let $\tilde s_k = \frac{s_k}{\sqrt{\lambda_k}}$. These variables have zero mean, unit variance, and are mutually uncorrelated. The probability-flow ODE preserves the marginals, so $\x_1 \sim \Normal(\0,\I)$. Since $\operatorname{Cov}(\x_1) = \I$, the best linear predictor of $\tilde s_k$ from $\x_1$ explains the variance
  \begin{equation}      
    r_k = \|\operatorname{Cov}(\x_1, \tilde s_k)\|^2 = \|\operatorname{Cov}(\drift(\x_\epsilon), \tilde s_k)\|^2,
  \end{equation}
  where the second equality follows from the independence of $\z$ and $\x_0$.

  Both $\x_1$ and $\z$ have zero mean, so the drift does as well. Its best linear predictor from the centred data is
  \begin{equation}
    \widehat{\drift} = \sum_k \operatorname{Cov}(\drift(\x_\epsilon), \tilde s_k) \tilde s_k.
  \end{equation}
  By orthonormality,
  \begin{equation}
    \mathbb{E}\|\widehat{\drift}\|^2 = \sum_k \|\operatorname{Cov}(\drift(\x_\epsilon), \tilde s_k)\|^2 = \sum_k r_k.
  \end{equation}
  The identity follows from the definition of $\eta$, and the inequality follows because linear prediction cannot explain more than the total drift energy.
\end{proof}

The proposition therefore connects linear recovery to two properties of the drift. First, its energy and second, how much of it is linearly related to the data. These two determine how well the data components can be recovered collectively. However, reconstruction quality also depends on which components are recovered. Recovering components that contain much of the image variance contributes more than recovering those that contain little. Based on this, we now examine when the drift-energy share is a useful proxy for learnability and when it can be unreliable.

\paragraph{Which components are recovered matters.}
The connection to the drift share is particularly simple when the data variance is spread equally across all $d$ directions. In this case, each component contributes the same amount to reconstruction, so $\mu_k=1/d$. Using \cref{prop:readable_directions} and $\energy{\x_1}=d$, we obtain
\begin{equation}
  \label{eq:readout_isotropic}
  R^2_{\mathrm{lin}}
  = \frac{1}{d}\sum_k r_k
  = \eta\,\frac{\energy{\drift(\x_\epsilon)}}{\energy{\x_1}}.
\end{equation}
Hence, for isotropic data, the fraction of image variance recovered by a linear predictor is exactly the drift-energy share multiplied by $\eta$, the fraction of drift energy linearly explained by the data.

For images, however, some components contain much more variance than others. Recovering these components improves reconstruction more than recovering components with little variance. The readout therefore depends not only on the drift energy, but also on which components remain readable through the noise. This is captured by the weights $\mu_k$ in \cref{eq:readout}. Directions with no data variance demonstrate this issue particularly clearly. They contribute neither to the covariance nor to $V$ in \cref{eq:readout}, so they leave $R^2_{\mathrm{lin}}$ unchanged, but they still count towards the ambient dimension $d$ and therefore lower the drift-energy share without changing learnability. The weights $\mu_k$ account for this distinction, whereas normalisation by $d$ alone does not. Therefore, for general data, the drift-energy share alone does not determine reconstruction quality. The same sum $\sum_k r_k$ can give different values of $R^2_{\mathrm{lin}}$, depending on whether the readable components contain much or little of the image variance. We therefore examine which components remain readable in our experiments.

\paragraph{What this means for closed-form couplings.}
Under the scaling approximation in \cref{eq:delta_approx}, \cref{prop:readable_directions} gives $\sum_k r_k=\mathcal{O}(\log N)$. In our experiments, $\energy{\drift(\x_\epsilon)}\approx 16$--$22$, which constrains linear recovery across components, but does not rule out substantial reconstruction if the readable components contain much of the image variance.

We examine which components remain readable using the directional correlation $\rho_k^2$ in \cref{tab:per_direction}, which gives a lower bound on $r_k$.
\begin{equation}
  \rho_k^2
  = \frac{\operatorname{Cov}(s_k, e_k)^2}{\lambda_k}
  \le r_k.
\end{equation}
For closed-form endpoints, only $3$--$6$ leading directions have $\rho_k^2>1/2$, and these hold $47$--$70\%$ of the data variance. Thus, strong directional readout is concentrated in a few leading components. The full linear predictor may also extract information from other endpoint coordinates, which is measured separately in \cref{tab:linear_readout}.

These observations support the initial signal-to-noise picture: the drift retains image structure, but some components are much easier to extract from the noisy endpoint than others. At lower resolution, the strongly readable directions hold a larger share of image variance, while the drift energy barely changes. This is consistent with the improved linear readout and student reconstruction in \cref{tab:linear_readout,tab:test_reconstruction}, and explains why their improvement need not follow the $\frac{1}{d}$ growth of the drift share.

\begin{table}[ht]
  \centering
  \footnotesize
  \caption{\textbf{Strong directional readability is concentrated in a few leading directions for closed-form couplings.} We report the number of principal directions with $\rho_k^2>\frac{1}{2}$ and the share of data variance they hold, for closed-form and teacher endpoints. This threshold selects directions in which a linear predictor using the corresponding endpoint coordinate recovers more than half of the component's variance.}
  \label{tab:per_direction}
  \begin{tabular}{llr rr rr}
    \toprule
     &  &  & \multicolumn{2}{c}{$\cpf$} & \multicolumn{2}{c}{$\cfmf$} \\
    \cmidrule(lr){4-5} \cmidrule(lr){6-7}
    dataset & px & $d$ & directions & variance & directions & variance \\
    \midrule
    CIFAR-10 & 16 & 768 & 5 & $58\%$ & 238 & $98\%$ \\
    CIFAR-10 & 24 & 1728 & 3 & $48\%$ & 174 & $94\%$ \\
    CIFAR-10 & 32 & 3072 & 3 & $47\%$ & 154 & $85\%$ \\
    \midrule
    CelebA & 16 & 768 & 6 & $70\%$ & 254 & $97\%$ \\
    CelebA & 32 & 3072 & 6 & $64\%$ & 425 & $89\%$ \\
    CelebA & 64 & 12288 & 6 & $61\%$ & 5262 & $84\%$ \\
    \midrule
    ImageNet & 16 & 768 & 6 & $61\%$ & 248 & $97\%$ \\
    ImageNet & 32 & 3072 & 6 & $56\%$ & 173 & $90\%$ \\
    ImageNet & 64 & 12288 & 6 & $53\%$ & 148 & $85\%$ \\
    \bottomrule
  \end{tabular}
\end{table}

\section{Fitting the training pairs and the random baseline}
\label{app:pair_fitting}

On ImageNet, the gap score $G$ is considerably higher than on CIFAR-10 and CelebA \crefp{fig:dimension_ladder}, although the closed-form couplings are no more informative there \crefp{tab:cosine,tab:linear_readout}. In this section, we show that the higher score is caused by the random baseline, which cannot fit its $1.28$M training pairs.

Every student is trained on a fixed set of $N$ pairs, and a student that learns its coupling well will also fit these pairs, i.e. map each training endpoint $\x_1^i$ close to its paired image $\x_0^i$. Fitting the training pairs is therefore not a sign of overfitting. For the teacher coupling, which is a smooth deterministic map from noise to data, fitting the pairs and generalising coincide, as the $\cfmf$ students reconstruct $92\%$–$99\%$ of the variance of held-out images from unseen endpoints \crefp{tab:test_reconstruction}. The situation is different for the random coupling. Since $\x_1$ is independent of $\x_0$, the optimal velocity at $t = 1$ is $\mathbb{E}[\x_1 - \x_0 \mid \x_1] = \x_1 - \bar\x$, so the optimal one-step student outputs the mean image $\bar\x$. A random student can only improve on the mean image by fitting its finite set of pairs, and its quality therefore depends on whether $N$ pairs can be fitted within the training budget.

\paragraph{Probing pair fitting.} We feed each student its own training endpoints and compare the one-step output with the paired image, which gives the training $R^2$ in \cref{tab:test_reconstruction}. On CIFAR-10 and CelebA, both the random and the closed-form students fit their pairs, with $R^2$ between $0.85$ and $0.97$. On ImageNet, with $80$ epochs over $1.28$M pairs, the random student fits its pairs only partially at $16$ px ($R^2 = 0.62$) and hardly at all at $32$ and $64$ px ($0.17$ and $0.02$), and it has no information in its coupling to fall back on. At $64$ px it is effectively a mean predictor, with an FID of $415$, close to the FID of the mean image ($446$). The closed-form student does not fit its pairs perfectly either, but it still recovers $65$--$82\%$ of the paired image's variance and $40$--$51\%$ on held-out images, consistent with it decoding the drift.

\paragraph{Control with 50k pairs.} To isolate the effect of $N$, we train the ImageNet-16 students on the first $50$k pairs of each coupling, with the same network, batch size and number of steps \crefp{tab:n50k_control}. The random student improves eightfold at one step and reaches the level of the CIFAR-10 random student, and $G$ drops to the CIFAR-10 level. The $\cfmf$ student becomes worse. Its coupling already contains the full noise-to-data map, so the student's quality is set by how densely the pairs sample this map, and $50$k pairs sample it $25$ times more sparsely than $1.28$M.

In conclusion, the high $G$ on ImageNet reflects a random baseline that cannot fit its pairs, not closed-form couplings that are better than on the other datasets. Since the random students are similarly poor at every ImageNet resolution, with one-step FIDs between $416$ and $430$ \crefp{tab:latent_fid}, the increase of $G$ as $d$ decreases remains meaningful within ImageNet.

\begin{table}[h]
\centering
\caption{\textbf{Controlling for the number of pairs.} ImageNet-16 students trained on all pairs ($80$ epochs) and on the first $50$k pairs ($2048$ epochs), with the same network, batch size and number of steps (best checkpoint).}
\label{tab:n50k_control}
\begin{tabular}{llcccccccc}
\toprule
 & & \multicolumn{2}{c}{$\crnd$} & \multicolumn{2}{c}{$\cpf$} & \multicolumn{2}{c}{$\cfmf$} & \multicolumn{2}{c}{$G$} \\
\cmidrule(lr){3-4} \cmidrule(lr){5-6} \cmidrule(lr){7-8} \cmidrule(lr){9-10}
dataset & pairs & 1 step & 2 steps & 1 step & 2 steps & 1 step & 2 steps & 1 step & 2 steps \\
\midrule
ImageNet-16 & $1.28$M & 430.1 & 158.0 & 97.5 & 44.2 & 2.3  & 1.6  & 0.78 & 0.73 \\
ImageNet-16 & $50$k   & 53.8  & 28.2  & 35.3 & 25.1 & 12.1 & 11.6 & 0.44 & 0.18 \\

\bottomrule
\end{tabular}
\end{table}

\section{Control experiments}
\label{app:controls}

The experiment in \cref{sec:latent} reduces the dimension by downsampling and keeps the network fixed across resolutions. This leaves possible confounders: downsampling changes the image content along with the dimension, a fixed network has more capacity per data dimension at low resolution, and at low resolution the teachers might memorise their training images. We address each of them in turn. Unless stated otherwise, all experiments follow the setup of \cref{sec:exp_detail}.

\subsection{Reducing the dimension with a learned autoencoder}
\label{app:autoencoder}

Downsampling removes fine image detail, which could change how hard the data is to learn independently of $d$. To test if our results depend on the dimension reduction technique, we therefore reduce the dimension in a different way. We use a learned autoencoder that preserves the image content with high fidelity, and repeats the full construction of \cref{sec:latent} in its latent space.

\paragraph{Setup.} On CIFAR-10, we train a KL-regularised autoencoder following \cite{rombach2022high} with an L1, a perceptual (VGG16) and an adversarial (PatchGAN) loss, and a KL weight of $10^{-5}$. The encoder downsamples the images by a factor of $4$ to $8 \times 8$ latents with $4$, $8$ or $16$ channels, i.e.\ $d \in \{256, 512, 1024\}$. The reconstruction FID of the training set is at most $0.95$ \crefp{tab:ae_ladder}, far below all student FIDs, so the autoencoder does not limit the comparison. In each latent space we train a flow-matching teacher, which reaches an FID of $7.4$, $6.1$ and $6.3$ for $d = 256$, $512$ and $1024$. The three coupling sets are constructed exactly as in pixel space. All FIDs are computed on decoded samples against the training images.

\paragraph{Results.} \Cref{tab:ae_ladder} shows the same trend as in pixel space. At one step, $G$ grows from $0.14$ at $d = 1024$ to $0.49$ at $d = 256$, and at two steps from $0.24$ to $0.39$. Since the autoencoder alters the data in a very different way than downsampling, the improvement of the closed-form couplings at lower dimension does not depend on how the dimension is reduced. The absolute values of $G$ are not comparable to those in pixel space, since the latent distribution differs from the image distribution. As discussed in \cref{app:learnability}, readability depends on how the data variance is distributed across directions and not on $d$ alone.

\begin{table}[ht]
  \centering
  \caption{\textbf{Dimension reduction with a learned autoencoder (CIFAR-10).} One- and two-step FID of the students trained on $\crnd$, $\cpf$ and $\cfmf$ in the latent space of the autoencoder (mean over three seeds), and the resulting gap score $G$ \crefp{eq:relative_score}. rFID is the FID of the autoencoder's reconstructions of the training set.}
  \label{tab:ae_ladder}
  \begin{tabular}{rcr rr rr rr rr}
    \toprule
     & & & \multicolumn{2}{c}{$\crnd$} & \multicolumn{2}{c}{$\cpf$} & \multicolumn{2}{c}{$\cfmf$} & \multicolumn{2}{c}{$G$} \\
    \cmidrule(lr){4-5} \cmidrule(lr){6-7} \cmidrule(lr){8-9} \cmidrule(lr){10-11}
    $d$ & latent & rFID & 1 step & 2 steps & 1 step & 2 steps & 1 step & 2 steps & 1 step & 2 steps \\
    \midrule
    256  & $4 \times 8 \times 8$  & 0.95 & 69.3 & 41.7 & 43.8 & 32.4 & 17.7 & 17.7 & 0.49 & 0.39 \\
    512  & $8 \times 8 \times 8$  & 0.55 & 71.1 & 40.3 & 49.4 & 34.9 & 20.8 & 20.9 & 0.43 & 0.28 \\
    1024 & $16 \times 8 \times 8$ & 0.37 & 71.2 & 56.0 & 64.8 & 48.2 & 24.1 & 23.1 & 0.14 & 0.24 \\
    \bottomrule
  \end{tabular}
\end{table}

\subsection{Equal capacity per dimension}
\label{app:capacity}

In \cref{sec:latent}, the students on CIFAR-10 use the same network at every resolution, so the number of parameters per data dimension grows from $18.4$k at $32$ px to $73.4$k at $16$ px. To remove this confounder, we reduce the base width of the network from $128$ to $96$ channels at $24$ px and to $64$ channels at $16$ px, which gives $18.4$k parameters per data dimension at every resolution, and retrain all three students on the same couplings. \Cref{tab:capacity} shows that the smaller networks leave the students on teacher couplings essentially unaffected, but make the random and closed-form students worse in absolute FID. The largest effect is on the random student at $16$ px, which relies on fitting its training pairs (\cref{app:pair_fitting}). Nevertheless, the gap score changes by at most $0.04$ ($0.01$ at two steps) and still doubles from $32$ to $16$ px. Lowering the dimension therefore improves the closed-form students also at equal capacity per dimension.

\begin{table}[ht]
  \centering
  \caption{\textbf{Students at equal capacity per dimension (CIFAR-10).} One-step FID of the students trained on $\crnd$, $\cpf$ and $\cfmf$ with the base width of the network either fixed ($128$) or scaled with $d$, and the resulting gap score $G$ \crefp{eq:relative_score}.}
  \label{tab:capacity}
  \begin{tabular}{rrrrr rrr r}
    \toprule
    px & $d$ & width & parameters & per dimension & $\crnd$ & $\cpf$ & $\cfmf$ & $G$ \\
    \midrule
    32 & 3072 & 128 & 56.4M & 18.4k & 85.1 & 69.2 & 14.7 & 0.23 \\
    \midrule
    24 & 1728 & 128 & 56.4M & 32.6k & 74.3 & 50.8 & 9.5 & 0.36 \\
    24 & 1728 & 96  & 31.7M & 18.4k & 84.8 & 58.5 & 9.1 & 0.35 \\
    \midrule
    16 & 768  & 128 & 56.4M & 73.4k & 62.9 & 33.0 & 4.7 & 0.51 \\
    16 & 768  & 64  & 14.1M & 18.4k & 93.9 & 52.1 & 4.7 & 0.47 \\
    \bottomrule
  \end{tabular}
\end{table}

\subsection{Memorisation}
\label{app:memorisation}

At $16$ px, the CIFAR-10 network has more parameters ($56.4$M) than the training set has pixel values ($38.4$M). A teacher that memorises its training images would, like the closed-form flow, send each noise to a training image. Its couplings would then trivially agree with the closed-form regions \crefp{app:coupling_comparison}, and the comparison between teacher and closed-form couplings would lose its meaning. We test for this in two ways.

\paragraph{Teacher samples are novel.} We generate 5000 samples from fresh noise with the 16 px teachers of CIFAR-10 and CelebA and compare each with its nearest training images. A copy of a training image would be much closer to it than to any other training image, so the ratio between the distances to the nearest and the second-nearest training image would be close to 0. Instead, the median ratio is 0.98 on CIFAR-10 and 0.97 on CelebA, and the samples are about as far from the training set as held-out test images. The students do not copy training images either, with median ratios of 0.97–0.98 on CIFAR-10 at 16 px.

\paragraph{Teacher couplings are insensitive to the injected noise.} Since the couplings are constructed in the direction from data to noise, we also test the teacher's map in this direction. A memorising teacher would have to spread the small perturbations around each training image over its entire region of prior space and would therefore be as sensitive to the injected noise as the closed-form flow. Repeating the noise-sensitivity analysis of \cref{tab:coupling_narrowness} at every resolution shows the opposite \crefp{tab:noise_share_ladder}. At most $8.2\%$ of the teacher's endpoint variance is due to the injected noise, compared with at least $97.7\%$ for the closed-form flow.

Hence, the teachers do not memorise their training images. In addition, the region agreement of \cref{app:coupling_comparison} is measured at the original resolutions, where the networks have far fewer parameters than the training set has pixel values ($0.37$ per pixel value on CIFAR-10). The agreement therefore reflects the consistency of the teacher with the closed-form partition and not memorisation.

\begin{table}[ht]
  \centering
  \caption{\textbf{Noise share across resolutions.} Fraction of the endpoint variance explained by the injected noise for teacher and closed-form couplings, measured as in \cref{tab:coupling_narrowness} on the first $2000$ training images with three independent noise draws ($\lambda = 1$, $\epsilon = 10^{-3}$).}
  \label{tab:noise_share_ladder}
  \begin{tabular}{lrr rr}
    \toprule
    dataset & px & $d$ & $\cfmf$ & $\cpf$ \\
    \midrule
    CIFAR-10 & 16 & 768   & $2.1\%$ & $98.2\%$ \\
    CIFAR-10 & 24 & 1728  & $2.7\%$ & $99.2\%$ \\
    CIFAR-10 & 32 & 3072  & $6.1\%$ & $99.6\%$ \\
    \midrule
    CelebA   & 16 & 768   & $1.2\%$ & $98.2\%$ \\
    CelebA   & 32 & 3072  & $3.8\%$ & $99.6\%$ \\
    CelebA   & 64 & 12288 & $8.2\%$ & $99.9\%$ \\
    \midrule
    ImageNet & 16 & 768   & $2.8\%$ & $97.7\%$ \\
    ImageNet & 32 & 3072  & $3.8\%$ & $99.4\%$ \\
    ImageNet & 64 & 12288 & $4.3\%$ & $99.8\%$ \\
    \bottomrule
  \end{tabular}
\end{table}

\end{document}